# Construction of Convex Sets on Quadrilateral Ordered Tiles or Graphs with Propagation Neighborhood Operations. Dales, Concavity Structures. Application to Gray Image Analysis of Human-Readable Shapes

Igor Polkovnikov AKA Igor Polk
*Yes San Francisco LLC, ipolk@virtuar.com, www.virtuar.com/ia/*
*December, 2015 - August, 2016*

## Abstract

An effort has been made to show mathematicians some new ideas applied to image analysis. Gray images are presented as tilings. Based on topological properties of the tiling, a number of gray convex hulls: maximal, minimal, and oriented ones are constructed and some are proved. They are constructed with only one operation. Two tilings are used in the Constraint and Allowance types of operations. New type of concavity described: a dale. All operations are parallel, possible to realize clock-less. Convexities define what is the background. They are treated as separate gray objects. There are multiple relations among them and their descendants. Via that, topological size of concavities is proposed. Constructed with the same type of operations, Rays and Angles in a tiling define possible spatial relations. Notions like "strokes" are defined through concavities. Unusual effects on levelized gray objects are shown. It is illustrated how alphabet and complex hieroglyphs can be described through concavities and their relations. A hypothesis of living organisms image analysis is proposed. A number of examples with symbols and a human face are calculated with new Asynchwave C++ software library.

## Keywords

# Introduction

This article gives more examples to the material described in the first article on this topic: "Asynchronous Cellular Operations on Gray Images Extracting Topographic Shape Features and Their Relations" published recently, see http://arxiv.org/abs/1303.4840. Also more functions are presented. The images of the first article are the colorful tables with the pixel values. It was done to simplify reviewing of the functions. These images were generated by the unique image analysis software and saved as HTML tables suitable for publication. This article presents images in a more familiar gray form. To review the pixel values, one should refer to the code. The code is made public and interested researchers may download it, review it, and analyze their own images with it. Search for "Asynchwave" software [3] at SourceForge.

Before going into the details, I'd like to stress the importance of the current research, the goals which were aimed.

Firstly, all these algorithms may be implemented in the parallel hardware working asynchronously. What it means. In the proposed hardware architecture, roughly speaking, each pixel depends on a certain min-max function of its neighbors. There is no synchronization needed. In other words, signals propagate in the hardware freely throughout the whole grid until the stable state is achieved. The signals propagate like a wave with the speed of light in the silicon and there is minimal power consumption since there is no clock.

Secondly, there are no arithmetic operations of addition or multiplication on the pixels. All operations are min-max operations with the exception of difference between images. Only a pixel and its immediate neighbors are used. The operations are asynchronous. This means that this simple kind of algorithms may be used by the living systems. I am not able to make any conclusions here, but I'd like to present this idea for the further research on how a living brain processes images. This is one of the ways possible.

Thirdly, I am presenting the way of describing structured patterns like symbols, characters, hieroglyphs, and signatures not as a set of strokes, but as a set of "dales" and "lakes". Gray dales on gray symbols. In my articles, I am talking about the whole system of pattern description with topological (or topographical if you wish) features, objects and their relations, which are supported by a large number of functions. The actual method of implementation which allows parallel asynchronous operation only supports the hypothesis that the "dales" are the primary features in the image processing by the brain. "Strokes" or "ridges" are the areas of objects bound by "dales". There is an example illustrating how to obtain them. Using the more abstract gray "dales" may greatly diminish dependence of the recognition algorithms on such variabilities as brightness, size, shape, stroke width, contour, angles, noise, uneven background, etc.

# Table of Contents

# 1. Basic Definitions

1.1. Assume that there is infinite quadrilateral tiling T. Each tile has 8 immediate neighboring quadrilateral tiles of two types, "+" and "x" connected. Any two "x"-connected tiles have a common vertex. Any two "+"-connected tiles have a common edge. There are four "+" and four "x"-connected neighbors of a tile.

1.2. 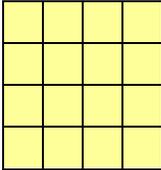 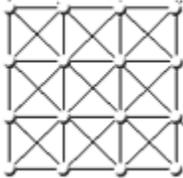 The T, shown on the left, is isomorphic to an infinite 8-regular graph with the structure shown on the right, when a tile corresponds to a vertex of the graph. Everything discussed here is related to such graphs either. The word "tile" can be replaced with the word "vertex". I am going to use the tile model, since it is more suitable for the applications discussed as well as to intuitive perception of the examples.

1.3. Though, the material presented does not count on any metric besides what is defined, it may be possible to use geometry in future, therefore the tiling model is preferable as far as I can see. Likewise graph theory may be used for further development, since the graph is a certain superposition of two meshes.

1.4. 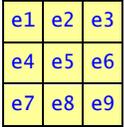 Let us abbreviate a tile as "e" or "e5" and its neighbors as on the Figure 1. The tiles e1, e3, e7, e9 are "x"-connected, e2, e4, e6, e8 are "+"-connected.

What properties of any pair of connected tiles can be observed:

1.5. A) Let us take any two "+"-connected tiles, say {e5,e2}. They have two symmetric pairs of 4 common neighbors, like {e1,e4},{e3,e6}. The pairs are not connected. Each tile within a pair is "+"-connected. Each of them is "x"-connected to one of {e5,e2} tiles and "+"-connected to another of {e5,e2} tiles.

1.6. B). Let us take any two "x"-connected tiles, say {e5, e1}. They have 2 common neighbors, i.e. {e2, e4}. They are "x"-connected among each other, and each of them is "+"-connected to each of the {e5, e1}.

1.7. A neighborhood of e5 can be designated N(e5). What properties of N(e5) can be observed:
C). In the N(e5), "+"-connected (to e5) tiles are "x"-connected among each other. For example e2 and e6; "x"-connected tiles (to e5) are not connected among each other. For example, e1 and e3;

1.8.

|    | e1   | e2   | e3   | e4   | e5   | e6   | e7   | e8   | e9   |
|----|------|------|------|------|------|------|------|------|------|
| e1 | self | +    |      | +    | x    |      |      |      |      |
| e2 | +    | self | +    | x    | +    | x    |      |      |      |
| e3 |      | +    | self |      | x    | +    |      |      |      |
| e4 | +    | x    |      | self | +    |      | +    | x    |      |
| e5 | x    | +    | x    | +    | self | +    | x    | +    | x    |
| e6 |      | x    | +    |      | +    | self |      | x    | +    |
| e7 |      |      |      | +    | x    |      | self | +    |      |
| e8 |      |      |      | x    | +    | x    | +    | self | +    |
| e9 |      |      |      |      | x    | +    |      | +    | self |

*Definition*: The tiling can be defined by the table of connectivity of any element (hence "e"), a tile or a vertex to its neighbors.

1.9. Let us abbreviate "+"-connected tiles as a neighborhood N4, "x"-connected tiles as Nx, and 8-connected as N8.

N4(e5)=={e2,e4,e6,e8}; Nx(e5)=={e1,e3,e7,e9}; N8(e5)=={e1,e2,e3,e4,e5,e6,e7,e8};

1.10. *Definition*: A **Path** P is a set of non-repeating connected tiles between any two it the tiling. For example, for N(e5), P(e1,e9)={e1,e2,e3,e6,e5,e8,e9}, P(e1,e5)={e1,e2,e5}, P(e1)={e1,e2,e4}. I use N(e5) only as an example. A path can be between any tiles in the tiling. Regarding connectedness, it can be P4, if all its tiles are "+" connected among each other, Px, if all its tiles are "x"-connected only among each other, and P8, if both types of connections is allowed;

1.11. *Definition*: A **Distance Path** DP is a shortest path between any two tiles. DP is measured in tiles. For example, DP(e1,e9)={e1,e5,e9}, DP(e1,e5)={e1,e5}, DP(e1)={e1}. A Distance Path may not be unique. For example, DP_1(e4, e6)={e4,e5,e6}; DP_2(e4,e6)={e4,e8,e6}. DP can be measured in the number of path tiles minus 1. Distances can be compared, for example, |DP(e2,e6)|>|DP(e1,e9)|. Again, there can be DP4, DPx, and DP8;

1.12. *Definition*: A **Distance Figure** DF is a union of Distance Paths between two tiles: DF8(e4,e6) = DP_1 ∪ DP_2 ∪ DP_3={e4,e2,e6}. There could be several kinds of DF.

Let us associate a tile with an ordered set V, so that every tile is associated with one element of the set V. For a graph it is called "to label". As an example, I will use the set of integers. We may say that "a tile e5 has value x", it is denoted as V(e5)==x, or just e5==x. But V does not have to be numbers as soon as relations <, <= are defined. Values of two or more tiles can be compared and their maximums or minimums found, for example, min(e5,e2). A tile can be assigned a value, like e5 = 5.

All tiles can be classified into subsets, called "colors". Sometimes it is convenient to denote one of such subsets as a "background" B. Than ~B is a "foreground" which comprises of all other subsets, and vice versa, if F is a foreground than ~F is a background. In the examples, the value of B tiles is assumed 0, but it can be any, if so specified.

One color may have a number of disconnected components.
*Definition*: A **Connected Component** is a subset of tiles for which there is a Path between every tile.

*Definition*: Paths and therefore Connected Components can be 1). **"+"-connected** (or "**4-connected**") if all tiles are "+"-connected and; 2). **"x"-connected** if all tiles are "x"-connected only, and 3). **8-connected** when all tiles may be connected in both/any ways. Abbreviations will be like DF4, or DFx, or DF8.

Here is an example. There are two colors, A and C. All tiles A are 4-connected, all tiles C are 4-connected. In this sense, there are two components here, AAA and CCC. The pair AAA and CCC is not 4-connected. It is 8-connected in a single component AAACCC. If "x"-connectivity only is considered in the A and C colors, there are three components here: AAC, CC, and A. The unmarked tiles are comprised of two components: an 8-connected large component and a single tile in the bottom-right corner.

The word "image" is used in the article denoting a visual representation of the tiling as a black and white photo. There is a number of examples in the article illustrated with images. Some of them are obtained with drawing in Photoshop or scanned. Other images are produced from the scanned images with the help of a software library "Asynchwave". The software applies the rules or devised algorithms to the input images and produces the result. A pixel "e" of an image corresponds to a tile or a vertex and its brightness or the "level of gray" corresponds to V(e). If viewed in electronic form the images can be enlarged and the content can be viewed in a greater detail.

Current imaging technology uses rectangular pixels and the article is about quadrilateral tiling, but I am trying to build a theory which is irrelevant to the shape and size of the tiles. Hopefully, the similar reasoning can be used to work on the tiles of arbitrary topology.

## 2. Problem 1: Construct Binary 4-Connected Convex Hull. Proofs

Assume, there is a large enough set of tiles colored into two components. One is B, another one is an 8-connected component S. Construct a union of all DF4-s among all tiles of this component.

2.1. The construction is done with an iterative process. Assume the accumulated constructed set is R (result). At first R=∅. Symbol ⊂ in the context of R means "the action of including a tile or a set of tiles into R". Note that if, at some stage, x ∉R, it does not mean that it may not belong after the next iteration. At the end, R will contain all the S tiles as well as some B tiles.

2.2. *Lemma:* if x∈R ⇒ x∈DP4. *Proof:* By construction, R is assembled from x∈DP4 ∎

2.3. 1). For any tile x belonging to the set S, x ⊂R, since any tile will belong to the final DF4.
∀x |∈ S ⇒ ∃ DP4 | x∈DF4 ⇒ S⊂R;

2.4.

2). Let us have a closer look how DP4 looks for two connected tiles {x,y}. For 8-connected components, there could be 2 cases. Firstly, if tiles {x,y} are "+"-connected, like {e5,e6}, DP4 includes both of them, the distance length is 1, |DP4|==1, i.e.
∀{x,y} | "+"-connected ⇒ {x,y} ∈DP4;

Secondly, if tiles are "x"-connected, DP4 should include a tile which is a"+"-connected neighbor of both {x,y}. |DP4|==2. There are two of DPs, for example for {e5,e9} the DP4 includes {e6,e8}. It means,
∀{x,y} | "x"-connected ⇒ their common neighbors | "+" connected ∈DP4;

It means for a "+"-connected pair, no new tiles are appended to DP4, for a "x"-connected pair, two new tiles are appended to DP4:
∀{x,y} | "x"-connected | ∈ S ⇒ their common neighbors | "+" connected ∈DP4;

2.5. 3) For a pair when {x,y} ∈ S, the rule 2) is applied, i.e.
if x and y are x-connected, than their common neighbors | "+" connected ⊂R;

2.6. 4). For the pair {x,y} | x ∈ S, y ∉S, y ∈ DP4, {x,y}∈R the rule is similar to 2.4:
∀{x,y} | "x"-connected ⇒ their neighbors | "+" connected ∈DP4 ⇒ their neighbors | "+" connected ⊂R;

*Proof*: Let us consider only one neighbor marked "?" on the Figure. Let us index the set NR of the immediate 8-connected neighbors"r" of all three tiles {x,y,?} with 1 to 12. The set NR surrounds {x,y,"?"} in such a way that neither {x,y,"?"} are connected to any other tile other than r∈NR.

Assume that y is on the shortest path "dp" between two tiles {rm,rn}∈NR and rm≠x, rn≠x, {rm,rn}∈R. (The case when y is on the shortest path between x and any r was considered in 2). ). What are rm and rn? If rm==r5, rn==r6, or {r5,r6}∈dp, than y ∉dp, since |DP(r5,r6)=={r5,y, r6}| >| DP(r5,r6)=={r5,r6}|. The same can be said about {r3,r5} and many other pairs. y ∈dp only if y ∈DP({r11,r12,r1,r2,r3}, {r6,r7,r8}) or DP({r3,r4,r5,r6}, {r8,r9,r10,r11}). For these sets, it can be shown that the set of shortest paths P(x, r) | r∈{rm, rn} also includes the tile "?". For example, let rm==r3, and rn==r6. Than DP(x, r6)=={x,"?", y, r6) or {x,r4,r5,r6} or ... .
Therefore, "?" ∈DP4, therefore "?"⊂R •

2.7. 5). For the pair {x,y} | {x,y} ∉S|{x,y}∈R the rule is similar to 2.6. 4): ∀{x,y} | "x"-connected ⇒ their neighbors | "+" connected ∈DP4 ⇒ their neighbors | "+" connected ⊂R.

*Proof*:
2.7.1. If r4∈R, than "?" ∈DP4, since |DP4(x,y)=={x,r4,y}| == |DP4(x,y)=={x,?,y}|, therefore "?"⊂R;
2.7.2. If r4∉R. It should be, but it is not in the R yet. Note, that "?" also in not in the R yet. Then we assume that x and y belong to different DPs. Since DP4 is 4-connected, than the only distance path which includes y and does not include {x,r4} is dp1={r6,y,r8}. Similarly, dp2={r2,x,r12}, i.e.

dp1 and dp2 belong to R already. Here we have found what r-s belong to R already, since it is given that {x,y} ∈R.
Let us take r12 and r6. No matter what is connected so far to r12, and to r6, i.e. what DP4(r12,r6) which does not include x, r4, "?", or y is, the distance |DP4(r12, r6)| if any, will be longer than |{r12,x,"?",y,r6}|. The same can be said about {r2,r6},{r12,r8},{r2,r8}.
Therefore, "?" ∈DP4, therefore "?"⊂R •

2.8.  *Rule 1*. Taking in account symmetry, then 2.3, 2.5, 2.6, and 2.7 can be summarized in one rule to solve the Problem 1 which is for ∀e5, if ({e2,e4} or {e2,e6} or {e6,e8} or {e4,e8}) ∈(S or R) ⇒ e5⊂R;
repeat;

It is applied to every tile until no tile value is changing. Thus the rule of construction of the Distance Figure==R is established. The computational process of construction of DF with the method described has the propagation nature, since it starts from the connected component spreading toward the "outlier" tiles.

2.9. So far nothing is said if this rule can find all possible tiles. Here is my attempt to prove it.

*Proof*:

2.9.1.  Let us take one tile t0∈S. Then by 2.3, t0⊂R.

2.9.2.  Then, let us take another t1∈S which is a neighbor of t0. There could be two cases. When t1 is "+"-connected, than t1∈DP4, DF consists of two tiles, all possible tiles are included into R.
When t1 is "x"-connected, than, as the result of rule, the R will accept 3 new members, since there are two DP from t0 to t1. The R now contains 4 tiles, all possible tiles are included. On the Figure, new non-S tiles are marked "+".

2.9.3.  Lets us take another t2 ∈S. There are three possible cases. After application of the rule, it is clear that all required tiles marked "+" are included into R.
*Note*. In the top illustration to 2.9.3, t2 is not "+"-connected to the original {t0,t1}∈S. It illustrates that tiles, which are "close enough" to S, also may be included into R. In practice, this can be desirable or not. If not, precautions should be taken to ensure that the rule is applied only to a connected component.

2.9.4.  Based on the previous, it can be shown, that for any arbitrary R, if one more tile is added and it is connected to the R tiles, after application of the rule, R will be amended without gaps. It proves that the application of the Rule 1 appends all possible tiles to R, and therefore R==DF •

2.10. One might say, that the Problem 1 is a problem of finding a convex hull on the connected components of the tile space. What would be a result of this rule or operation for complex shapes? Here is an example to obtain intuitive feeling. Note, that the Difference sets are not connected !

2.10.1. Original three connected sets:   2.10.2. Result of the operation:   2.10.3. Difference:

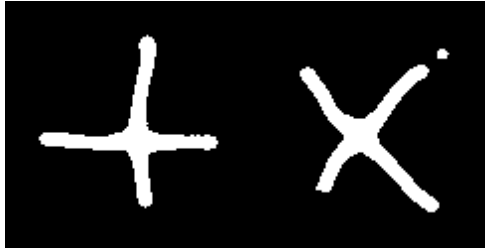 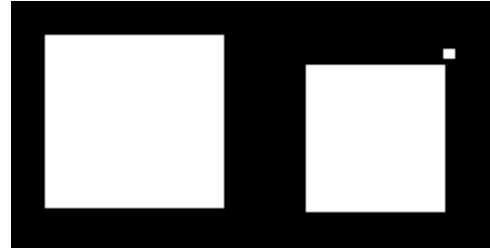 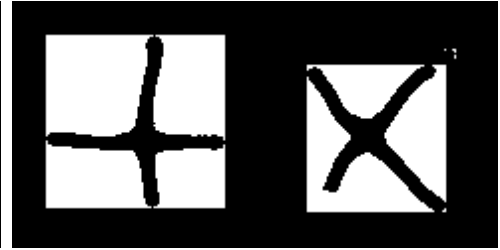

# 3. Problem 2: Construct Gray 4-Connected Convex Hull

3.0. Assume, there is a large enough set of tiles. The tiles are not colored, there are no disconnected components, there is no B and S partitioning, 8-connection among all tiles is assumed. Values on which the order is defined are assigned to each tile: $\forall e5 == v | \in V$. There is enough of tiles with v==0 (or in other words, e5==0) surrounding tiles e5>0. It does not have to be 0, any constant. What would be the meaning of **DF4** in this case?

3.1. The rule "**convex_hull4**" which is used to constructs DF4 is this:

for $\forall e5$, if ( e5<min(e2,e4) or e5<min(e2,e6) or e5<min(e6,e8) or e5<min(e4,e8) ) ⇒ e5 = max ( min(e2,e4),min(e2,e6), min(e6,e8)),min(e4,e8) ) repeat;

It is applied to every pixel until no pixel is changing. Note, that the same tile can change its value many times until the stability is reached.

3.2. After 3.1 the difference is found between the result of 3.1 and the original. Here is an example of the operation. In image analysis, the result may be called "gray convex hull". The original image was drawn in Photoshop(TM) with a brush. It has 256 levels of brightness and is very smooth, even though it looks like it has large flat regions. The resulting images are obtained with the public software "Asynchwave".

| 3.2.1. Original two connected sets: | 3.2.2. Result of the convex_hull4: | 3.2.3. Difference between 3.2.2 and 3.2.1: | 3.2.4. XOR of 3.2.2 and 3.2.1: |
|---|---|---|---|
| 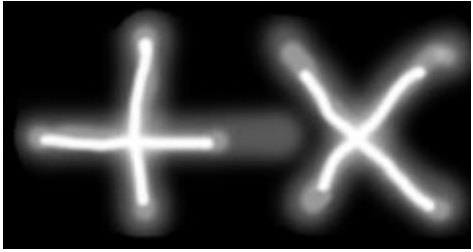 | 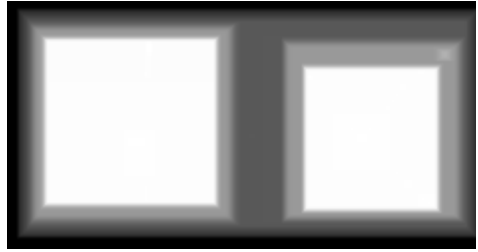 | 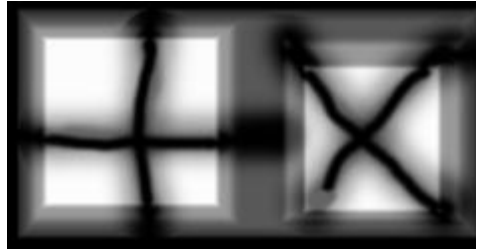 | 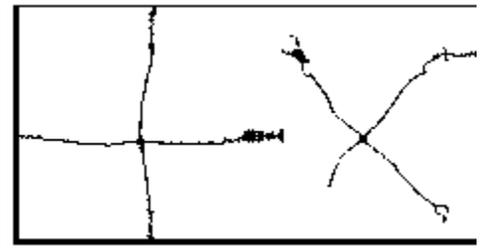 |

Note, that no background B value has been chosen. However the e5==0 has a certain significance. At 3.2.3 and 3.2.4, this value is possessed by tiles with the same value in original set and the result set. These tiles are black, v==0. They are not changed during the operation, while all tiles with values>0 were changed.

3.3. *Interpretation of the results*. In the original image 3.2.1, there are two distinct types of pixels due to the method of drawing. The first type is the pixels in the white areas of the crosses. Since they were drawn with the same brush, they have statistically the same value. Some of the pixels have exactly the same value and this value is the maximal of all pixels in the image. After the operation, paths among these pixels form the white flat DF area in the Result 3.2.2. This area covers some pixels which have the smaller value. These pixels can be identified in the "XOR of B and A", 3.2.4 figure as disruptures of the black irregular lines. The second type of pixels in the original image are at the diluted ends and the gray levels of the crosses. They look more similar to the images which could be obtained with a video camera. Black pixels in the "XOR of B and A" figure are the pixels which were not changed during the operation. I encourage you to look at all these images via a software tool with which you can visually evaluate the levels, like thresholding or "levels" in Photoshop(TM). All white pixels in the "XOR of B and A" figure represent DF4, black irregular lines belong to DF4 also, but they are unchanged. A thin black rectangular "frame" around, at the very border of the image 3.2.4 is the pixels which do not belong to the DF4, this is what someone can call the "background".

Therefore, on an "uncolored" tiling, DF is defined by the specific operation **convex_hull4** as a difference bounded by a precise rectangle. The same operation is used to define what is B: it is a value outside the rectangle. The **convex_hull4** partitions the tiling into components.

It is yet to be investigated, if the XOR always contains more than one connected white component. If so, it may serve as a characteristic of the original image. Each of these components and their pairs may be analyzed for specific properties.

In 3.1 the max-min function was used. It would be interesting to see an example of the min-min function. It may not give a stable or reproducible result.

# 4. Problem 3: Gray 8-Connected Convex Hulls

Assume, there is a large enough set of tiles. Values on which the order is defined are assigned to each tile: $\forall e5 == v | \in V$. Construct the **DF8**-s in the same manner as DF4.

4.1. The construction is done with an iterative process. Assume the constructed set accumulated is R (result). At first $R = \emptyset$. Symbol $\subset$ in the context of R means "the action of including a set of tiles or a tile into R". Note that if $x \notin R$, it does not mean that it may not belong after the next iteration.

4.2. *Lemma:* if $x \in R \Rightarrow \exists\ DP8\ |\ x \in DP8$. *Proof:* By construction, R is assembled from $x \in$ some DP8 •

4.4. 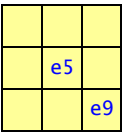 2). Let us have a closer look how DP is constructed for any two connected tiles $\{x,y\}$. For 8-connected components, there could be 2 cases.
Firstly, if tiles $\{x,y\}$ are "+"-connected, like $\{e5,e6\}$, DP8 includes both of them, the distance length is 1, DP value is 1.
$\forall \{x,y\}\ |\ $"+"-connected $\Rightarrow \{x,y\} \in DP8(x,y)$;

Secondly, if tiles are "x"-connected, DP8 does not necessarily includes a tile which is an "+"-connected neighbor of both $\{x,y\}$ since DP value is 1, and for the "+"-connected tiles it is be 2.
$\forall \{x,y\}\ |\ $"x"-connected $\Rightarrow \{x,y\} \in DP8(x,y)$, but not their "+"-connected neighbors;

4.5. 3) No new tiles should be appended to the R, for a pair of the tiles described in 4.4.

4.6. 4). What are the conditions when a new tile should be appended to the R?

4.6.1.

| | | e3 |
|---|---|---|
| | e5 | |
| | | e9 |

For any two {x,y} arranged as {e3,e9}, we can not add an "x"-connected component e5 arranged as on this figure, because it is not known that e3 and e9 belong to the same connected component, if connected components are to consider. Therefore it is required that in the e5 neighborhood it must be ensured that e3 and e9 belong to the same connected component. There could be several options.

| | | e3 |
|---|---|---|
| | e5 | e6 |
| | | e9 |

For any three {x,y,z} arranged as {e3, e6, e9} in the example, e5 belongs to DP8 since DP({e3,e6,e9})==2 and DP({e3,e5,e9})==2.

4.6.2.

| | e2 | |
|---|---|---|
| | e5 | e6 |
| | e8 | |

For any three {x,y,z} arranged as {e2,e6,e8} in the example, e5 belongs to DP8 since DP({e2,e6,e8})==2 and DP({e2,e5,e8})==2.

4.6.3.

| | e2 | |
|---|---|---|
| | e5 | e6 |
| | | e9 |

For any three {x,y,z} arranged as {e2,e6,e9} in the example, e5 belong to DP8 since DP({e2,e6,e9})==2 and DP({e2,e5,e9})==2.

4.6.4.

| e1 | e2 | |
|---|---|---|
| | e5 | e6 |
| | | e9 |

This combination is included into 4.6.3

4.6.5.

| | e2 | |
|---|---|---|
| e4 | e5 | e6 |
| | | e9 |

This combination is included into 4.6.2 or 4.6.3

4.7. *Summary*. All three cases, including symmetrical variations, must be included in the DF8 calculation. However, if there is a 4-connected component to consider instead of 8-, than the first case is sufficient, since out of all possible neighborhoods, only 4.6.1 is 4-connected. Here is a table of possible combinations:

4.7.1.

4.7.2.

4.7.3.

4.8. The general formula, "**convex_hull8on8**" is rather large. Avoiding mistakes, I am presenting it as a verified C-code (symbols || mean logical "or"):

d11 = min (e1,e2,e3);
d12 = min (e3,e6,e9);
d13 = min (e7,e8,e9);
d14 = min (e1,e4,e7);

d21 = min (e2,e6,e8);
d22 = min (e4,e2,e6);
d23 = min (e4,e8,e6);
d24 = min (e2,e4,e8);

d31 = min (e2,e6,e9);
d32 = min (e3,e6,e8);
d33 = min (e2,e3,e4);
d34 = min (e1,e2,e6);
d36 = min (e5,e6,e7);
d37 = min (e4,e8,e9);
d38 = min (e2,e4,e7);
d39 = min (e1,e4,e8);

if (e5<d11 || e5<d12 || e5<d13 || e5<d14 ||
 e5<d21 || e5<d22 || e5<d23 || e5<d24 ||
 e5<d31 || e5<d32 || e5<d33 || e5<d34 || e5<d36 || e5<d37 || e5<d38 || e5<d39 ){
   dm1 = max(d11,d12,d13,d14);
   dm2 = max(d21,d22,d23,d24);
   dm3 = max (d31,d32,d33,d34,d36,d37,d38,d39);
   dm = max (dm1,dm2,dm3);
   e5= dm;
}

4.9. Here is an example of an image in 256 levels:

4.9.1. A. Original connected set: 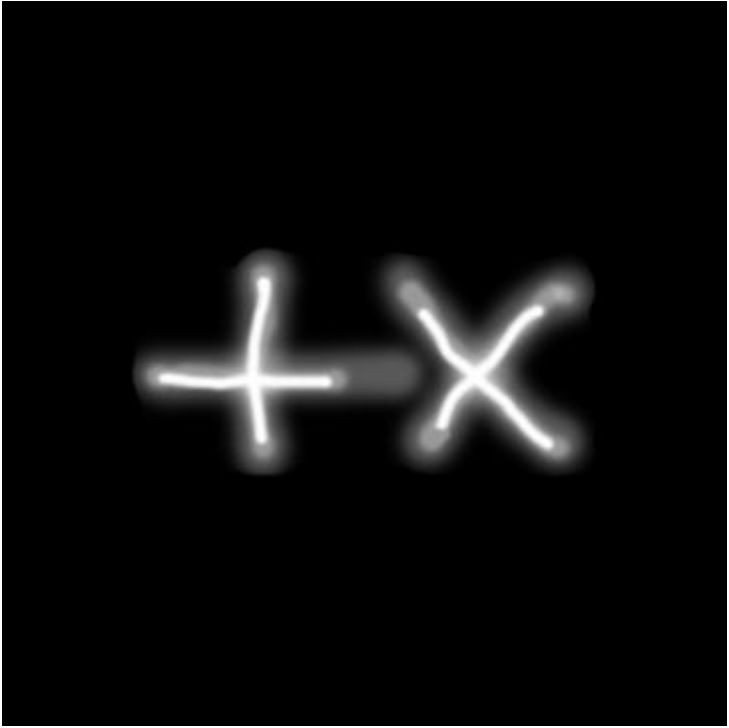

4.9.2. B. Result of the operation convex_hull8on8: 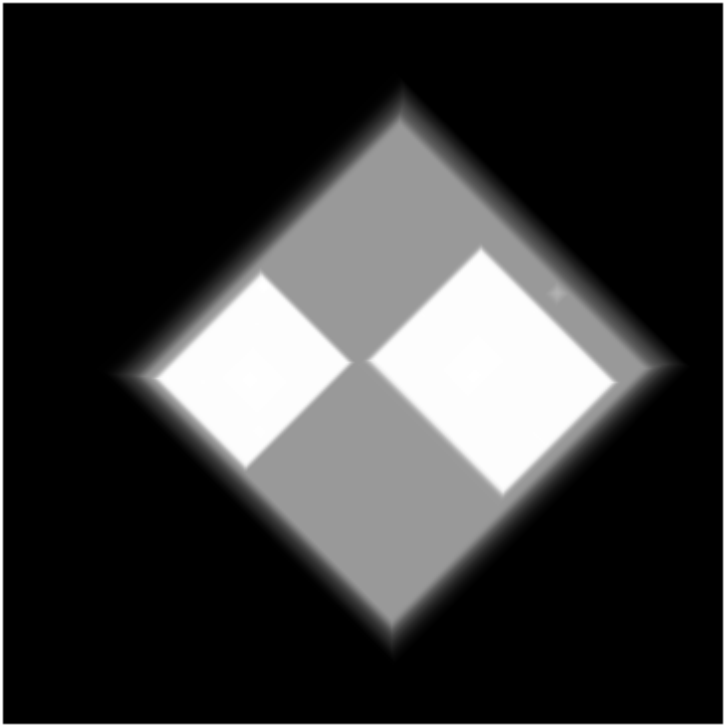

### 4.9.3. Difference between B and A:

### 4.9.4. XOR of B and A:

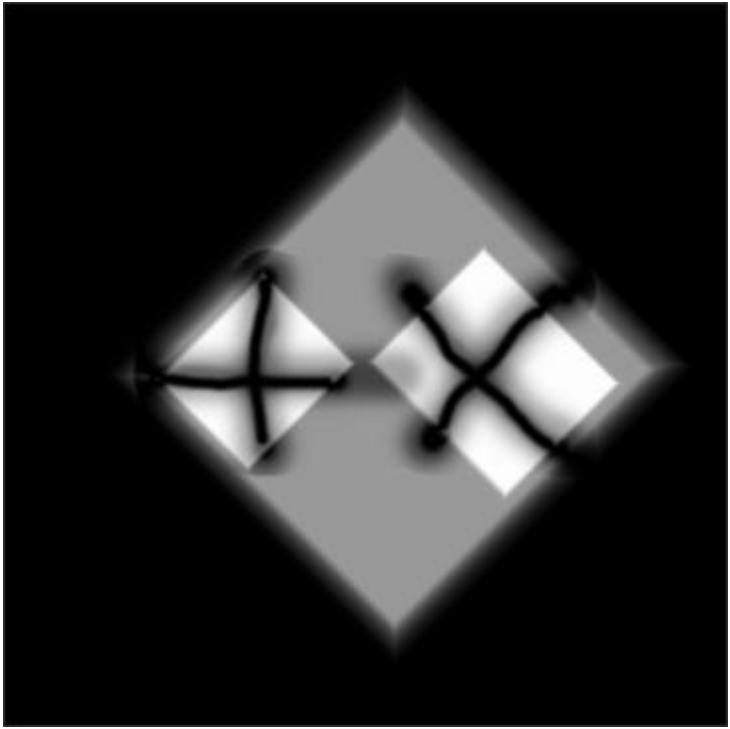
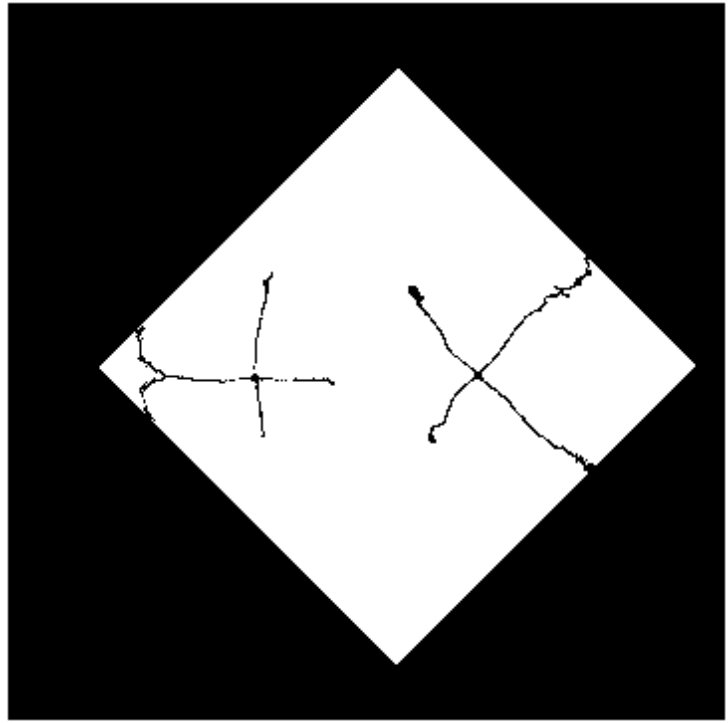

4.9.5. Interpretation of the result is similar to 3.3.

4.10. If we take only the 4.7.1 part of the formula, the result is practically the same. It is due to the original set is large and smooth. If the set would consist of individual tiles, or thread-like pattern the result might be different and complete formula should be used.

Here is the example below of the formula 4.7.1, **"expand_123_369_789_147"** and application of it to the same image.

```
d11 = min (e1,e2,e3);
d12 = min (e3,e6,e9);
d13 = min (e7,e8,e9);
d14 = min (e1,e4,e7);

if (e5<d11 || e5<d12 || e5<d13 || e5<d14){
  dm1 = max(d11,d12,d13,d14);
  e5= dm1;
}
```

| 4.10.1. A. Original connected set: | 4.10.2. B. Result of the operation expand_123_369_789_147: |
|---|---|
| 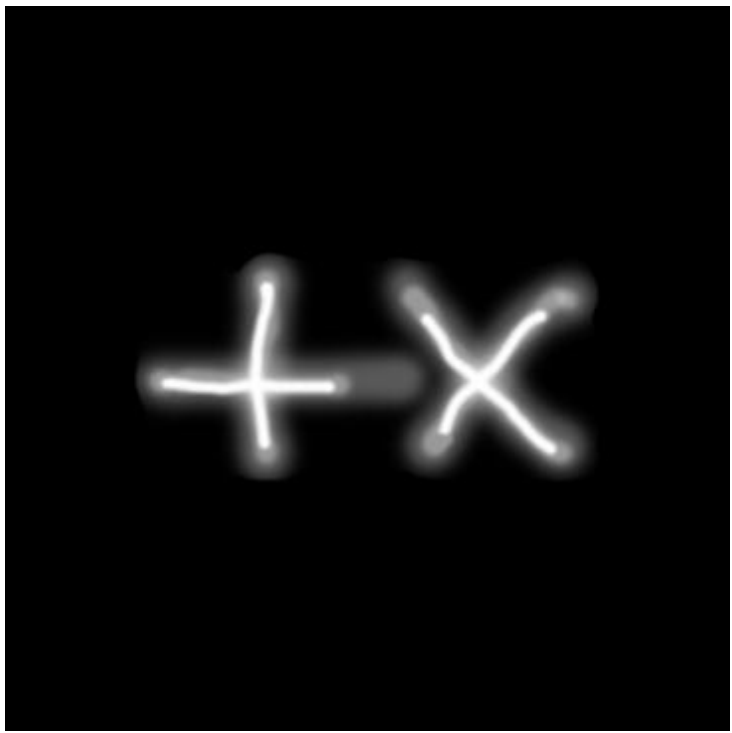 | 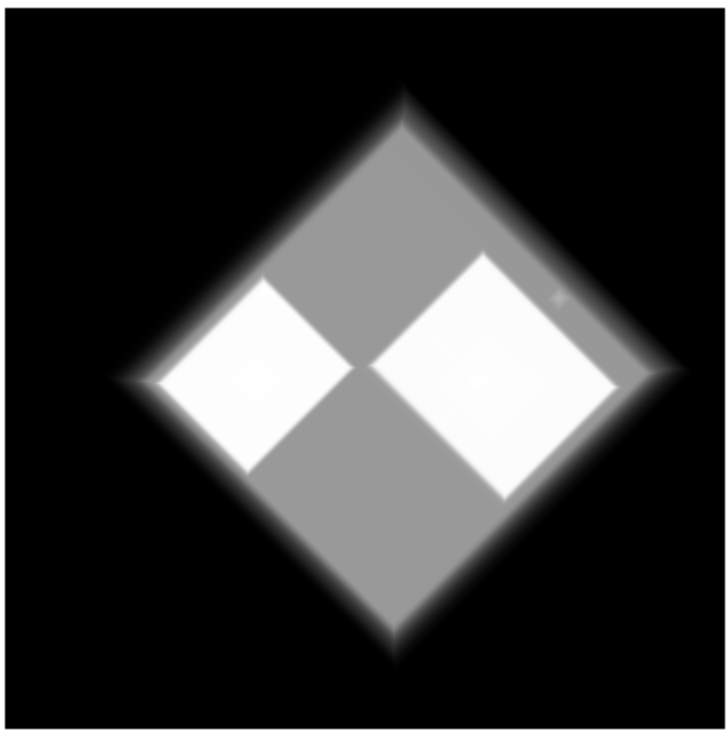 |

4.10.3. Difference between B and A:   4.10.4. XOR of B and A:

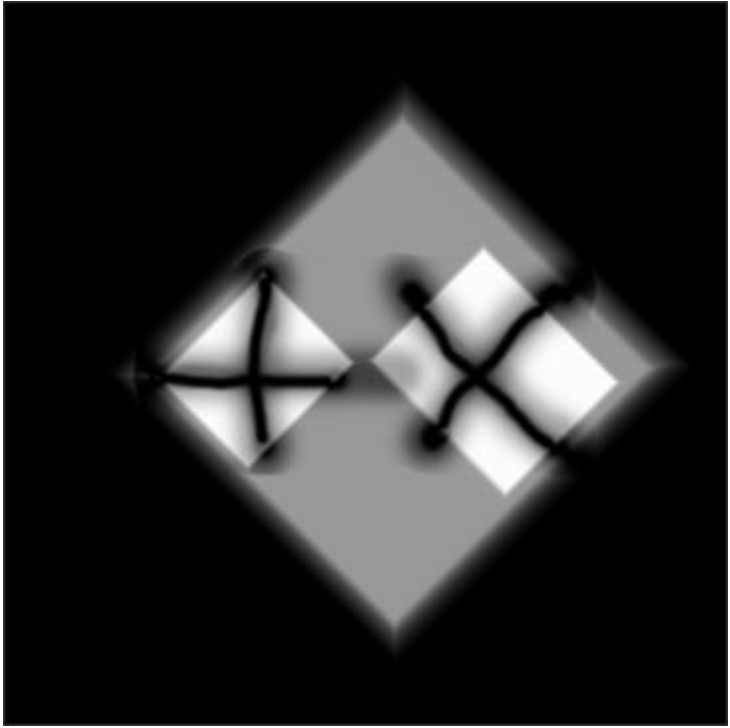 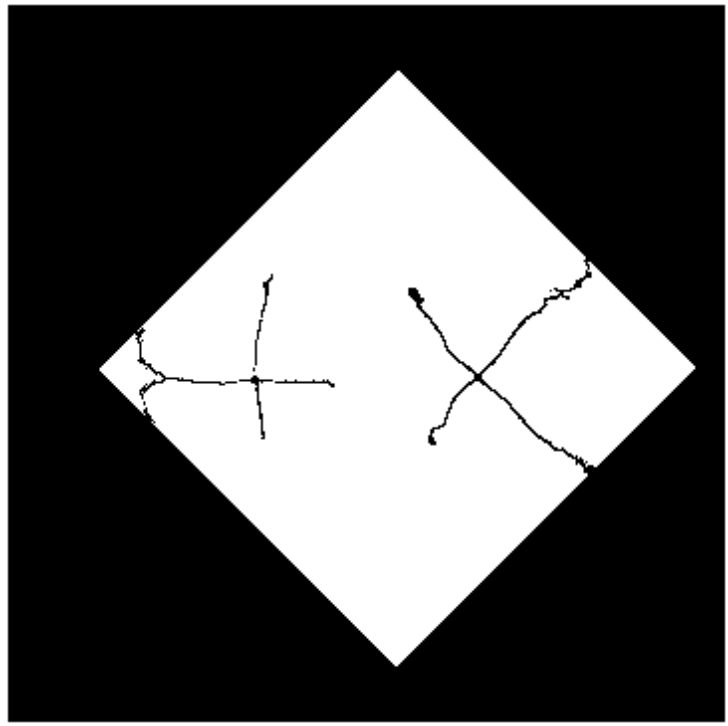

If you magnify the "XOR" images you will see the slight difference in pixels.

4.11. Here is the same image but in 4 levels processed with the same short formula **"expand_123_369_789_147"**:

4.11.1. A. Original connected set:

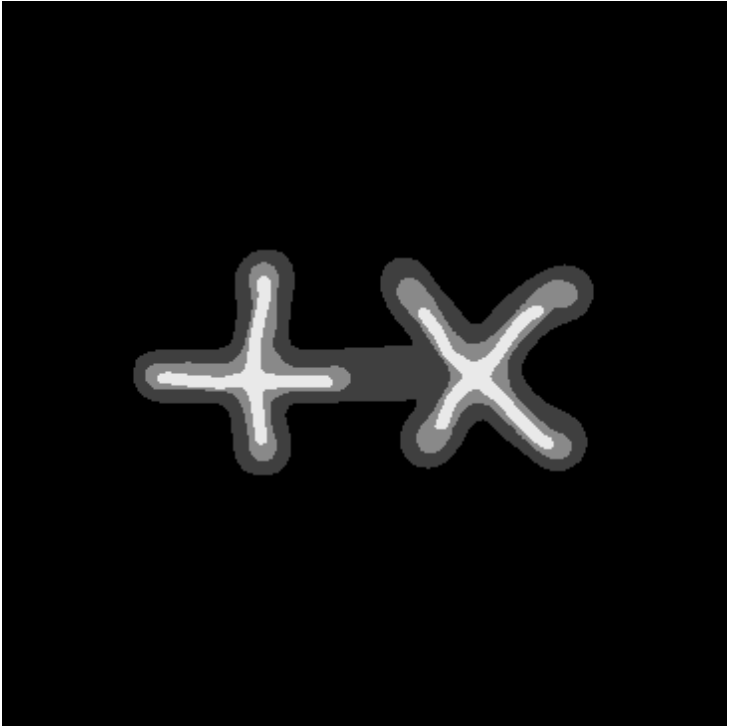

4.11.2. B. Result of the operation expand_123_369_789_147:

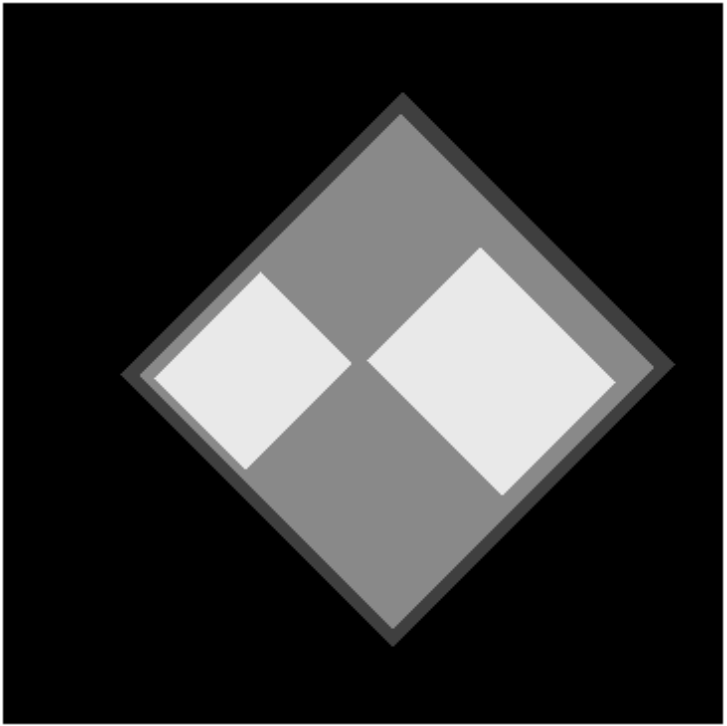

4.11.3. Difference between B and A:

4.11.4. XOR of B and A:

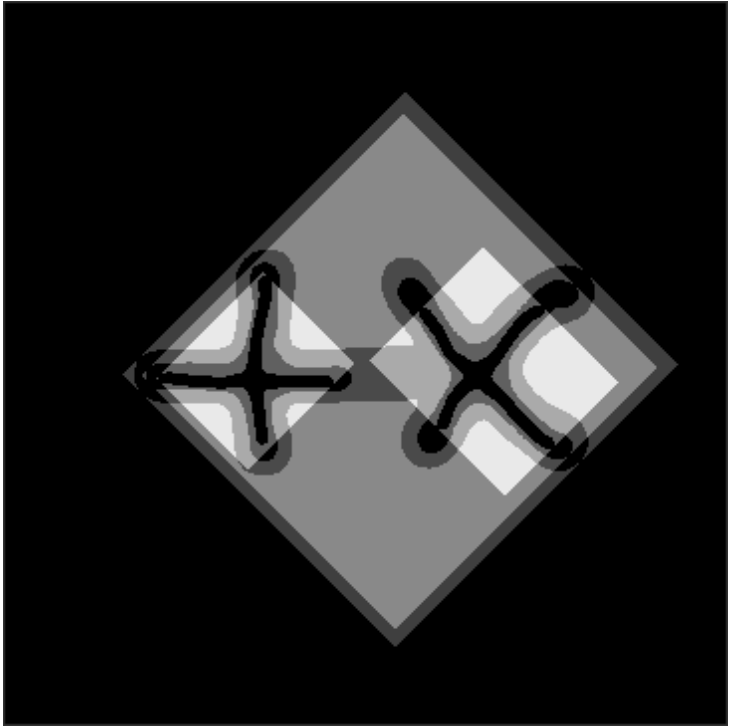
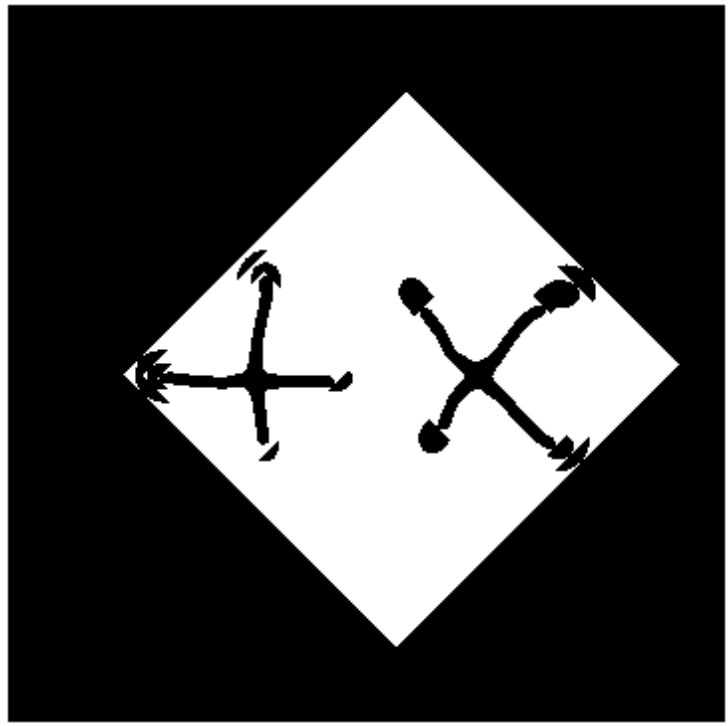

Here you can observe some saw-like artifacts which appear during operations on "levelized" images which are not smooth enough. One has to be aware of it. If the complete reverse of the **convex_hull8on8** - the **minconvex_hull8on8** (not present in the paper) were used, the result would be the same. Note, the black disconnected sets on the 4.11.4 at the ends of the crosses.

4.12. Here is an example with results of the operations to observe the importance of the 4.7.2 rule. As an original image, the same image as above with Gaussian noise superimposed was used:

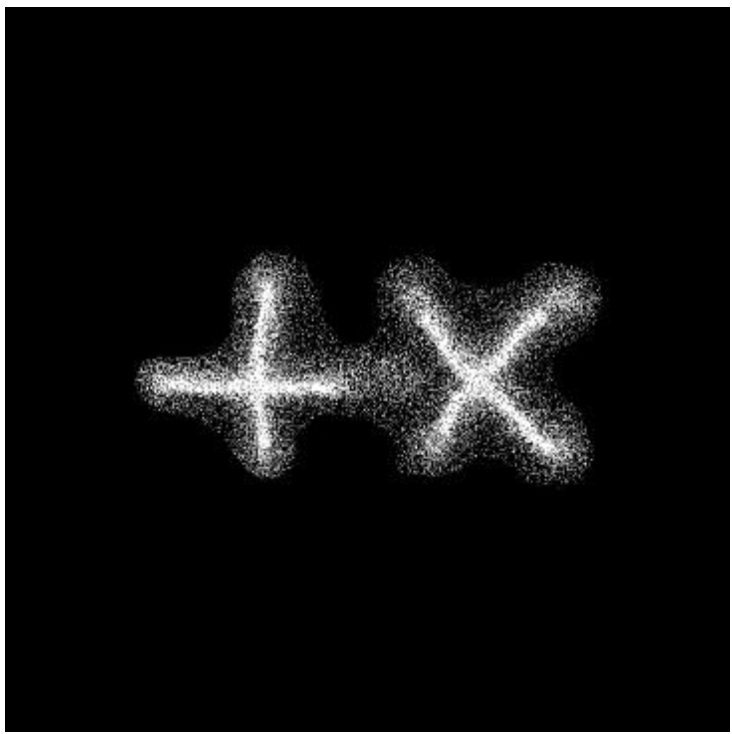

4.12.1. Complete rule 4.8, "convex_hull8on8":

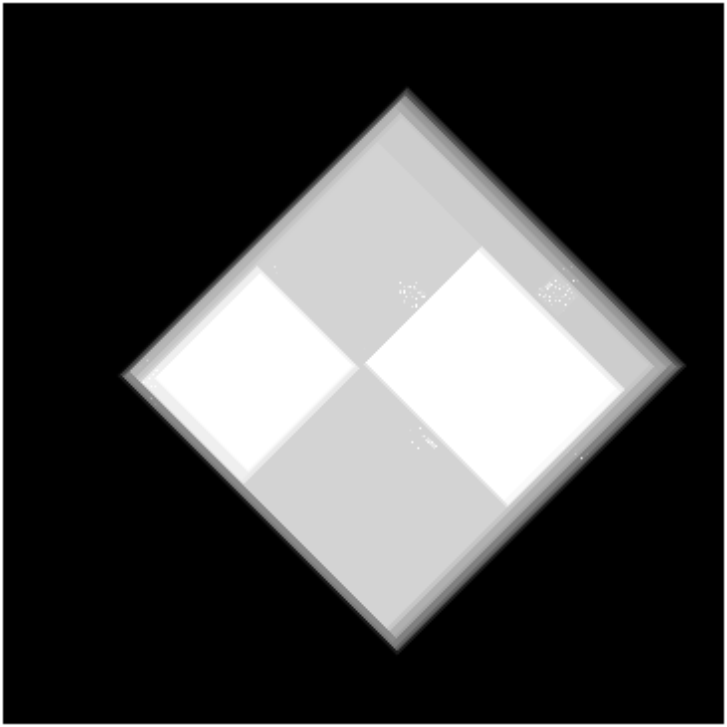

4.12.2. Application of the 4.7.1 rule only, "expand_123_369_789_147":

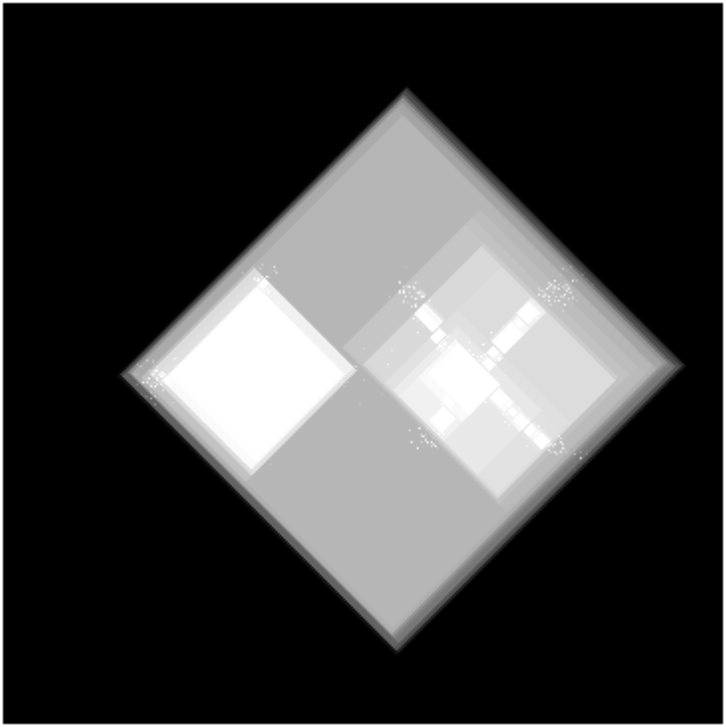

4.12.3. Application of the 4.7.2 rule only "expand_268_246_468_248": 4.12.4. Application of 4.7.1 and 4.7.2 rules together, "expand_268_246_468_248_123_369_789_147":

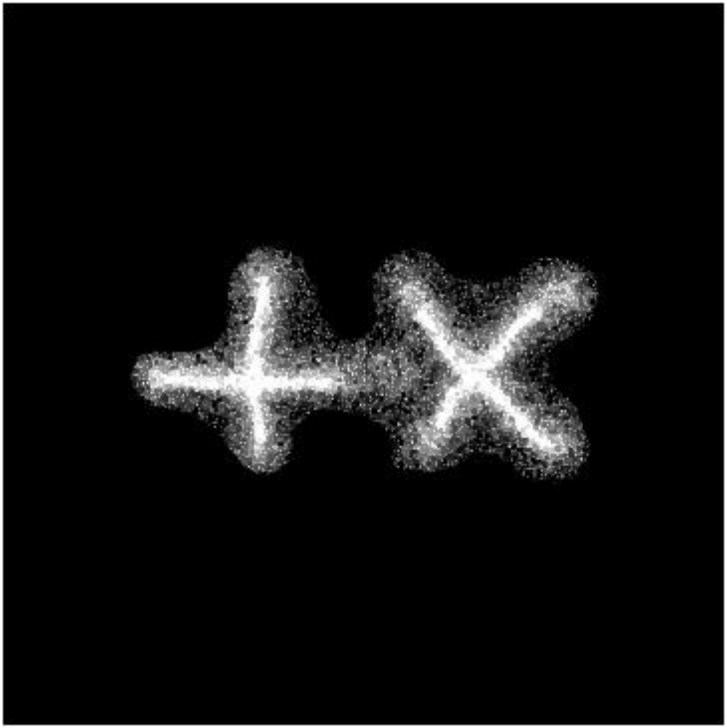 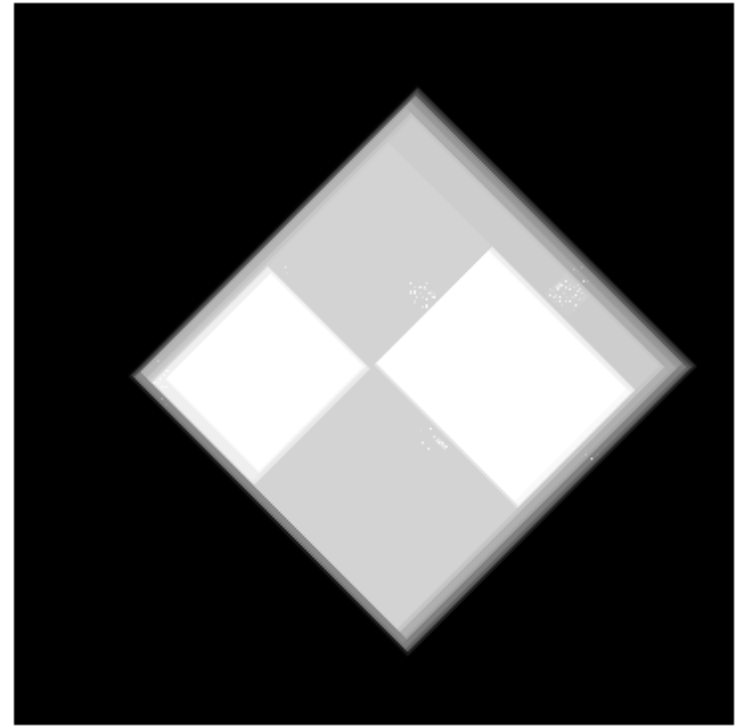

4.13. The rule 4.7.2 is important, if the tiling or expected components it operates upon are considered to be 8-connected. Without this rule, especially for thread-like patterns, the DF might be incorrect. However, for 4-connectedness of tiling or expected components, this rule is not necessary.

4.14. Now I'd like to present a surprising result of application of the 4.7.3 rule combined with the 3.1 rule in the following way. It is called "**convex_hull**" (symbols || mean logical "or"):

d1 = min (e4,e1,e2,e3);
d2 = min (e1,e2,e3,e6);
d3 = min (e2,e3,e6,e9);
d4 = min (e3,e6,e9,e8);
d6 = min (e6,e9,e8,e7);
d7 = min (e9,e8,e7,e4);
d8 = min (e8,e7,e4,e1);
d9 = min (e7,e4,e1,e2);

if (e5<d1 || e5<d2 || e5<d3 || e5<d4 || e5<d6 || e5<d7 || e5<d8 || e5<d9 ){
e5= max (d1,d2,d3,d4,d6,d7,d8,d9);
}

4.15.

| 4.15.1. | 4.15.2. | 4.15.3. | 4.15.4. |
|---|---|---|---|
| A. Original connected set: | B. Result of the operation "convex_hull": | Difference between B and A: | XOR of B and A: |

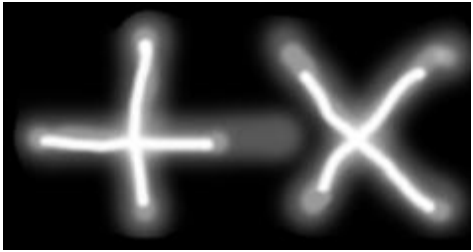 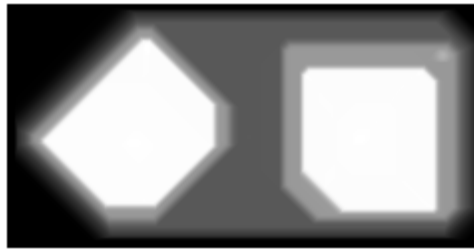 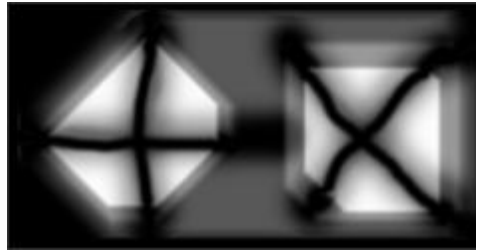 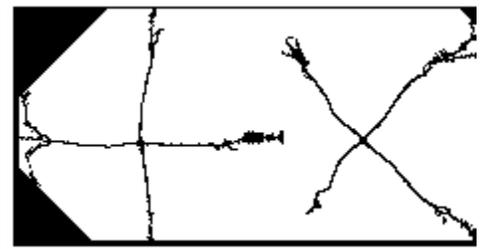

4.16. Another example made with a 4-level image:

4.16.1. A. Original two connected sets:

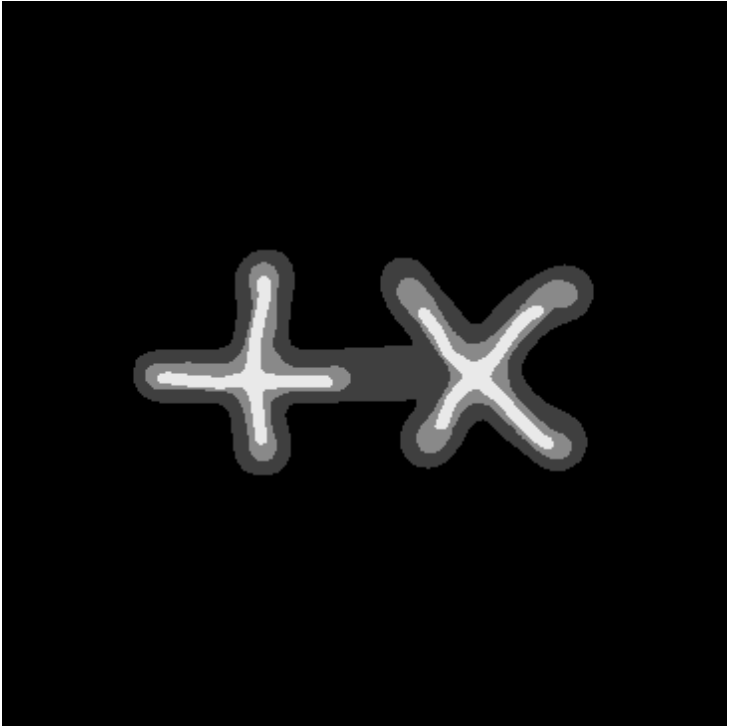

4.16.2. B. Result of the operation "convex_hull":

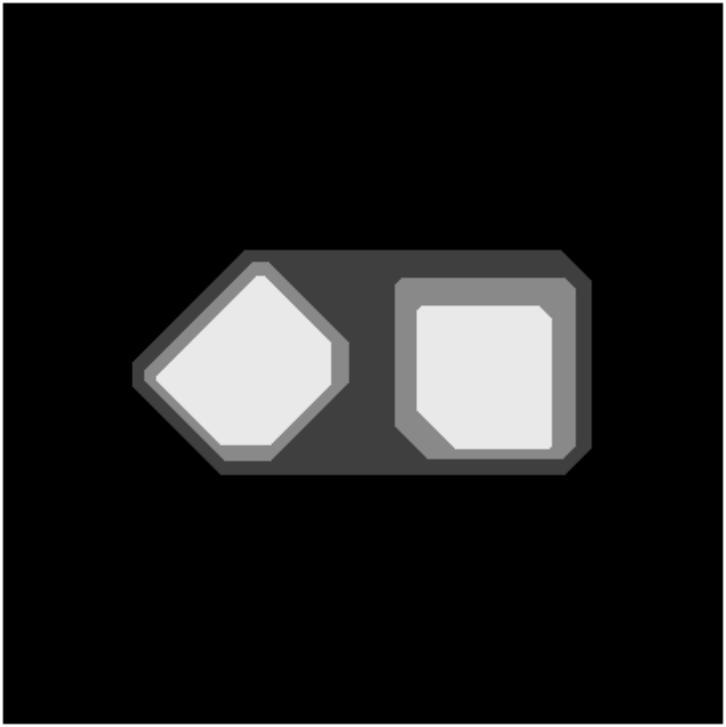

4.16.3. Difference between B and A:

4.16.4. XOR of B and A:

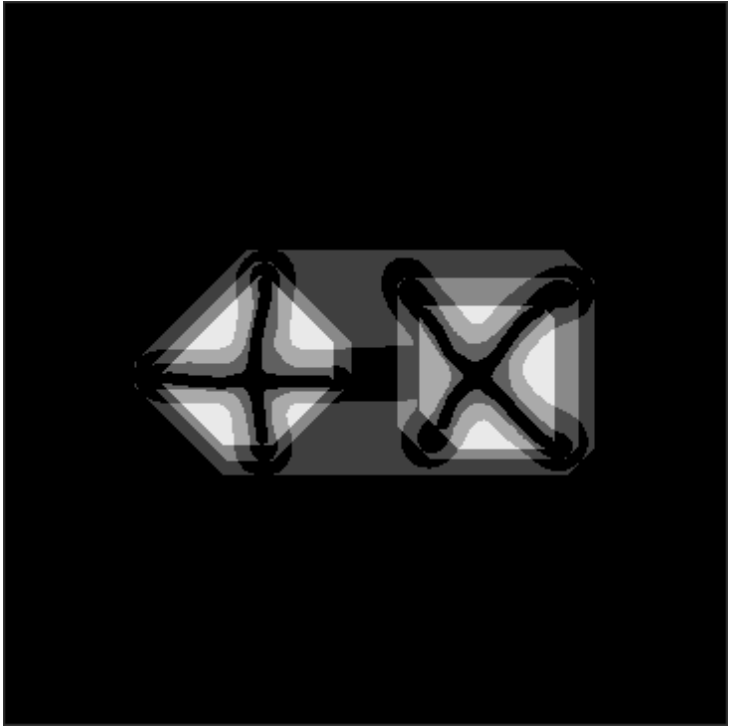
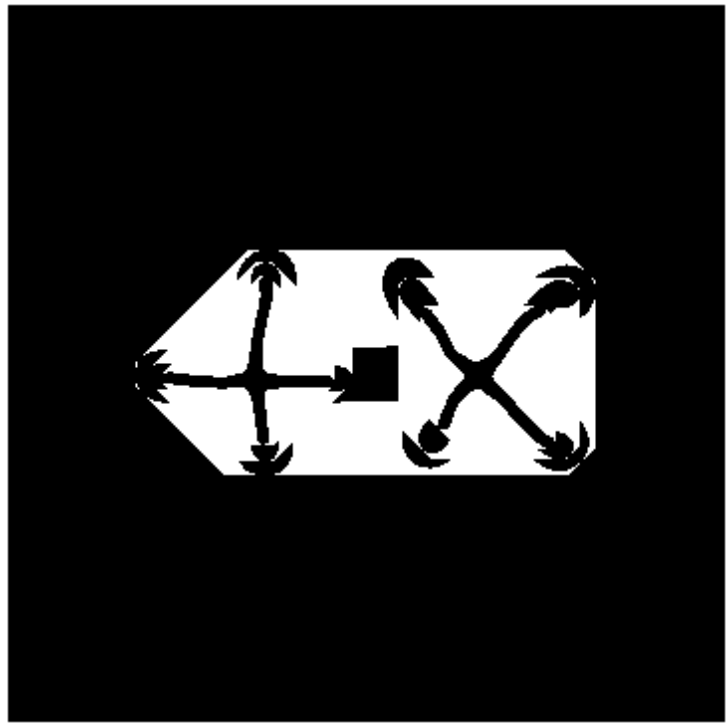

Currently, I am not sure, if this rule is complete while working with 4- or 8-connected images. It means, I do not know if some tiles (pixels) will be left out. But I have some confidence that it works well on 4-connected sets.

## 5. Stack of Tiling. Operations of Constraint Allowing Convex Hull Minimization

5.1. Assume there is a number of tilings, at least two, T1 and T2, stacked, T1→T2, T2→T1. Each tile of tiling T1 corresponds to a tile in T2 in such a way that its neighbors correspond to the neighbors of a tile ∈T2, or N(eT1)→N(eT2). The formula T2=T1 means that ∀e|∈T2 = e|∈T1, i.e. it designates the coping of values. We may consider connectedness, colors, and tile values of T1 and T2 separately. Let us introduce operations f(T1, T2). For simplicity, I will abbreviate e5 as one of the T1 tiles, and t5 as a corresponding T2 tile, e5∈T1, t5∈T2.

5.2. On stacks it is possible to make operations of "convex hull minimization". This is a "constraint" type of operations when values of one tiling define operation area in another tiling In tiling, since there are multiple DPs, it is possible to build convex hull of maximum area and minimum area. The following operations essentially invert the space and build convex hulls not of a connected component, but of the background surrounding it. Naturally, it may fill all the tiling, but the reference T2 prevents the process of entering deep into the original component. Mathematics of it will hopefully be presented in future papers.

5.2.1. Here is one operation "**minconvex_hull4**". It is done on the T1, using T2 as a reference only. It works as if T2 defines the region of allowable changes in T1. Note that this region is dynamic: T2 does not change, but T1 yes. The operation continues until no more tiles of T1 are due to change. The operation "**minconvex_hull4**" is equivalent to finding 4-connected convex hull "**convex_hull4**". The example of application is 5.3.3.

```
if ( e5 > t5 ){
  d24 = max (e2,e4);
  d48 = max (e4,e8);
  d68 = max (e6,e8);
  d26 = max (e2,e6);
  if (e5>d24 || e5>d48 || e5>d68 || e5>d26 ){
    e5= min (d24,d48,d68,d26);
}}
```

5.2.2 Here is the operation "**minconvex_123_369_789_147**" which is equivalent to "**expand_123_369_789_147**". The example of application is 5.3.5.

```
if ( e5 > t5 ){
  d123 = max (e1,e2,e3);
  d369 = max (e3,e6,e9);
  d789 = max (e7,e8,e9);
  d147 = max (e1,e4,e7);
  if (e5>d123 || e5>d369 || e5>d789 || e5>d147 ){
    e5= min (d123,d369,d789,d147);
}}
```

5.2.3. Here is the operation "**minconvex_hull**" which is equivalent to "**convex_hull**". The example of application is 5.3.7.

```
if ( e5 > t5 ){
  d1 = max (e4,e1,e2,e3);
  d2 = max (e1,e2,e3,e6);
  d3 = max (e2,e3,e6,e9);
  d4 = max (e3,e6,e9,e8);
  d6 = max (e6,e9,e8,e7);
  d7 = max (e9,e8,e7,e4);
  d8 = max (e8,e7,e4,e1);
  d9 = max (e7,e4,e1,e2);
  if (e5>d1 || e5>d2 || e5>d3 || e5>d4 || e5>d6 || e5>d7 || e5>d8 || e5>d9 ){
    e5= min (d1,d2,d3,d4,d6,d7,d8,d9);
}}
```

5.3. If T1 is an image of a convex hull, and T2 is an original image, the minconvex_hull4 decreases the size of the filled convexity.

5.3.1. Original connected set:

5.3.2. Result of the operation "convex_hull":

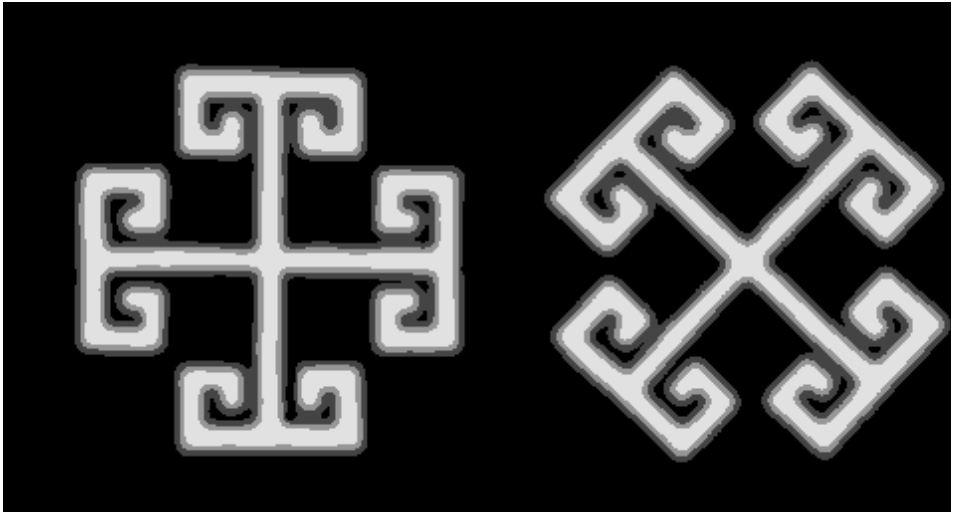
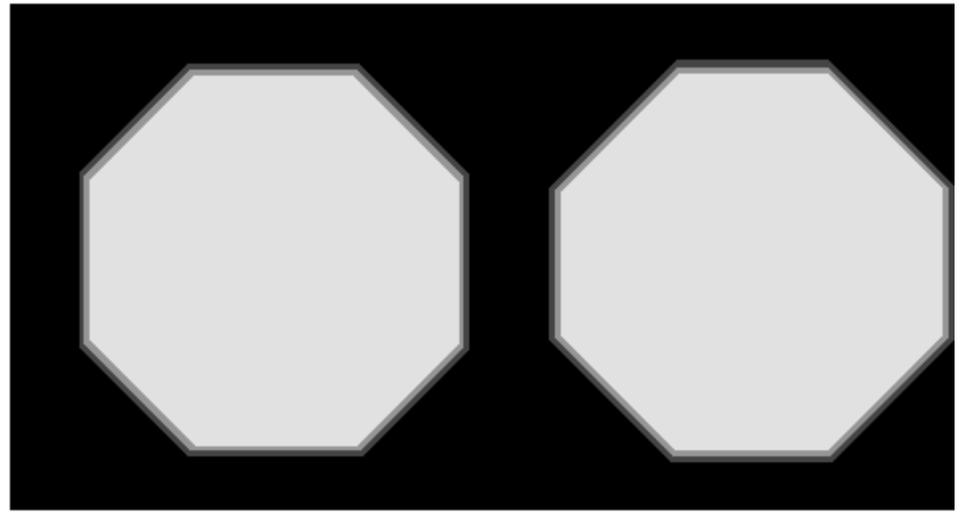

5.3.3. Result of "minconvex_hull4", 5.3.2 is T1, 5.3.1 is T2:

5.3.4 Difference of 5.3.3 and 5.3.1.

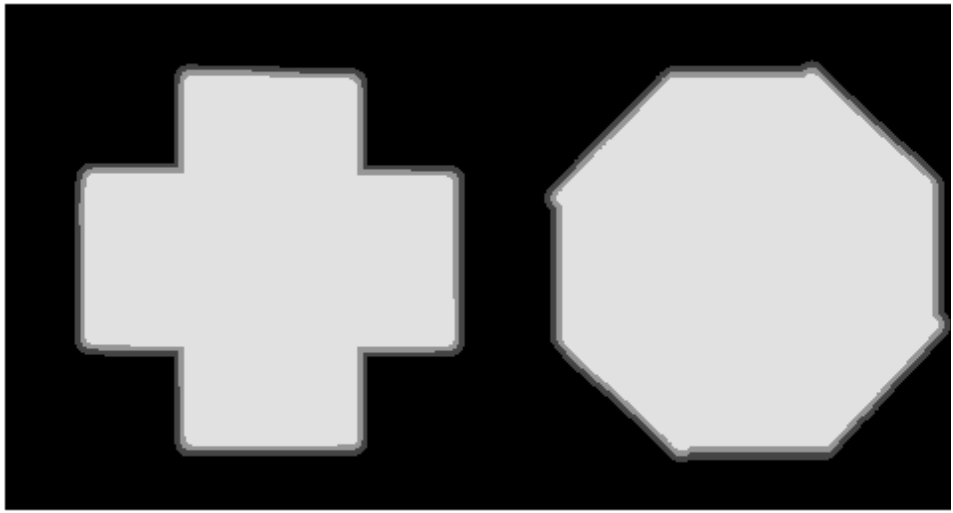
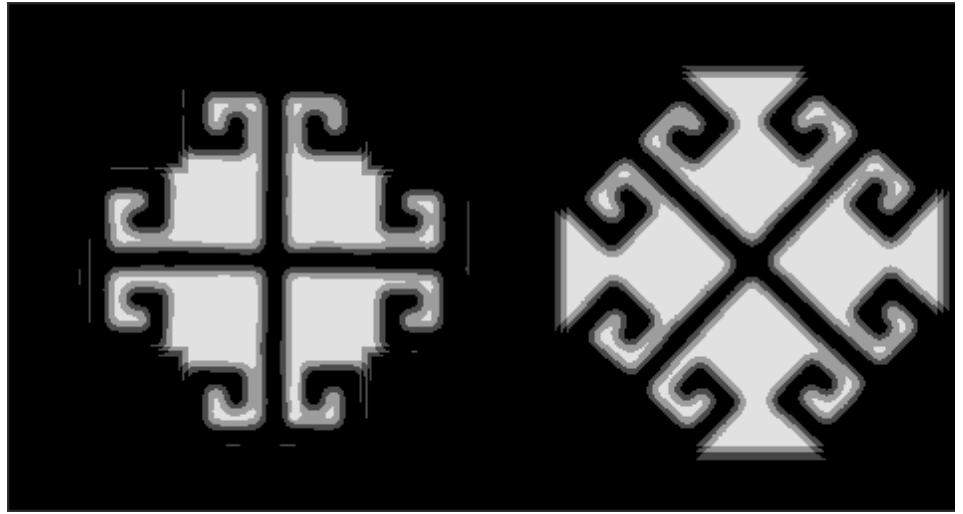

5.3.5. Result of "minconvex_123_369_789_147", 5.3.2 is T1, 5.3.1 is T2:   5.3.6 Difference of 5.3.5 and 5.3.1.

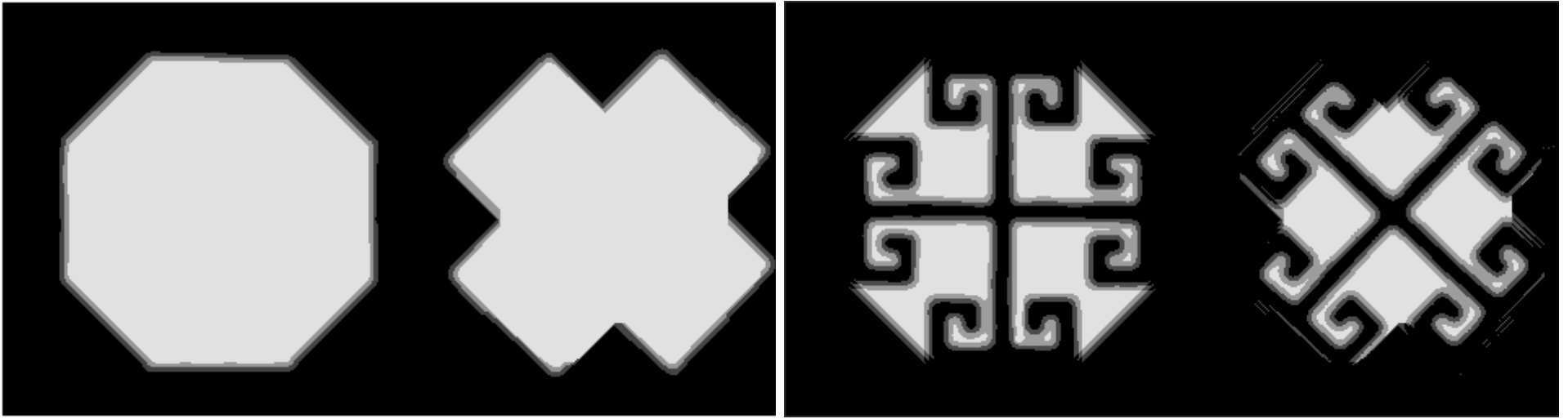

5.3.6.2. Here is the result of the complete reverse of the **convex_hull8on8** - the **minconvex_hull8on8** (not present in the paper). It is slightly different in details from the approximate function minconvex_123_369_789_147. The minconvex_hull8on8 is much more computationally intense than minconvex_123_369_789_147. How important is the difference, I can not say yet.

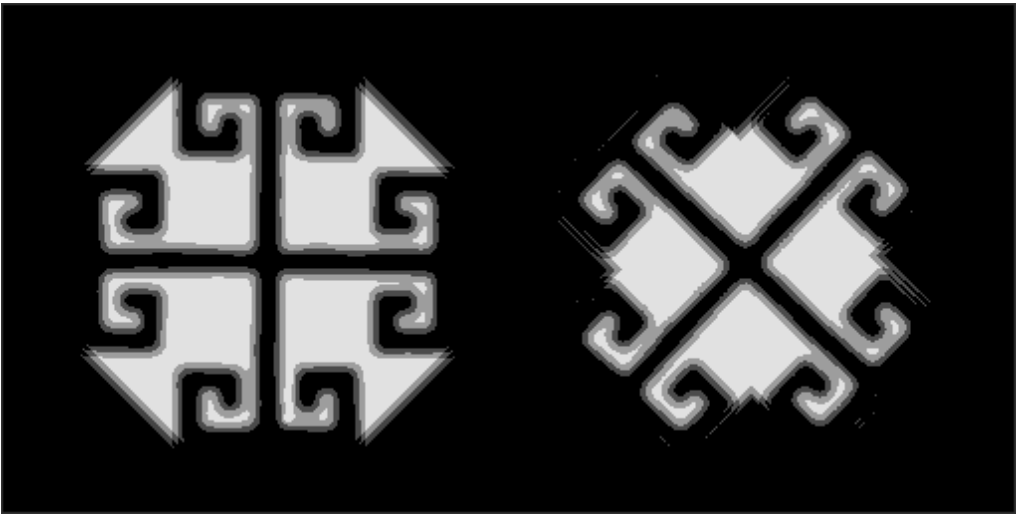

5.3.7. Result of "minconvex_hull". I use the original image 4.16.2, it is more informative in this case.

Difference

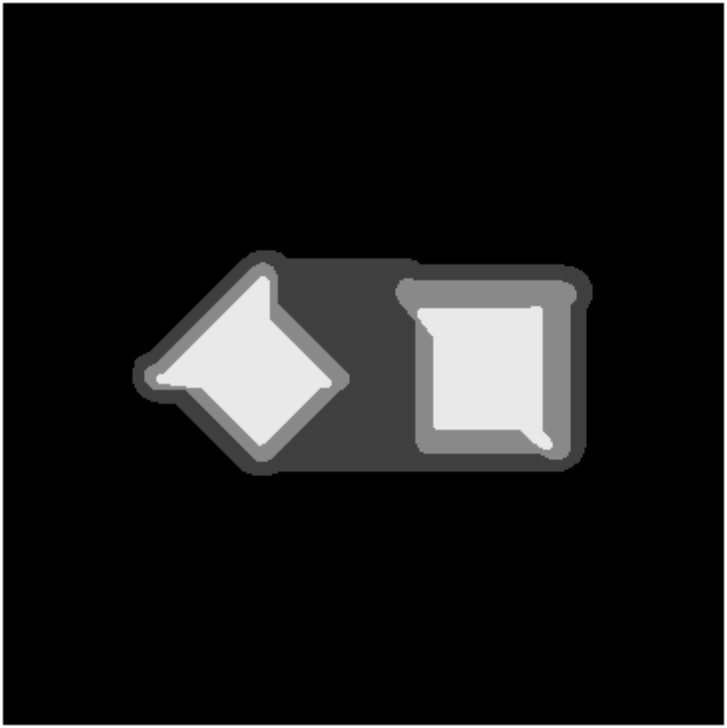
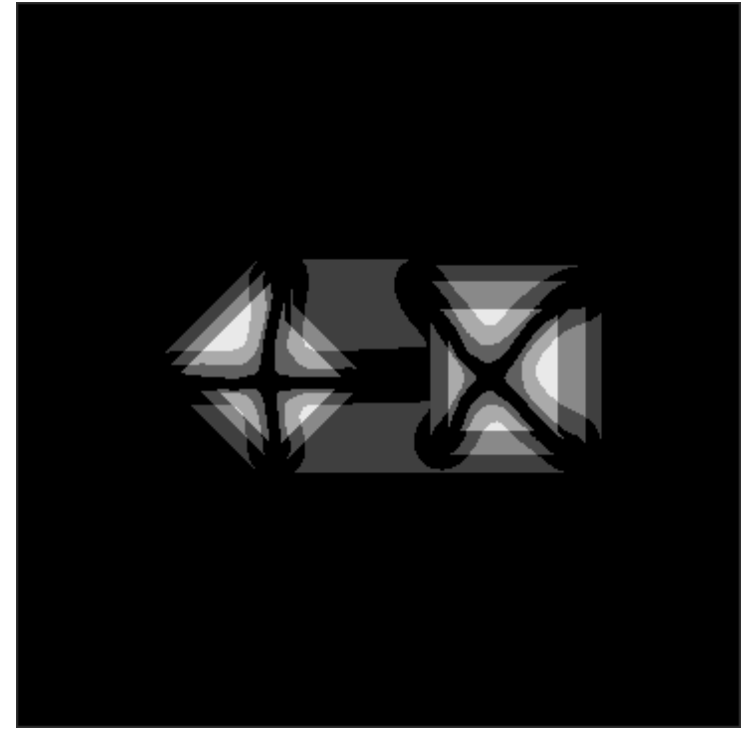

5.6. Functions **minconvex_hull4** and **minconvex_123_369_789_147** can be combined into a single operation, however this will not lead to the result which can be expected from application of the same functions consequently. The result will be the empty, exhausted set - not a convex hull: Fcombined ≠ F1 o F2.

## 6. Concavities, Minimal and Maximal. Measuring Their Size

6.1. The results of the mentioned above operations in 4. and 5. may be called "**concavities**". The application of the operations when one tiling contains a set and another one is its "convex hull" converts "maximal convex hull" to the "minimal convex hull" in the discreet space. Minimal convex hull has minimal area.

6.2. If the same operation, say 4-connected "**minconvex_hull4**" is repeated several times on the same tiling, there will be no new results. However, if the first operation is 4-connected and the following operation is 8-connected, the "concavities become deeper" and so forth, interchangeably: deeper and deeper, until the original convex hull is left with no concavities. In this case, some areas of the ordinal image may keep to be covered by the convex hull. These areas are "lakes". The lakes can be obtained by a single function, see later. The iterative process of the concavity "excavation" gives an idea that the size of concavities may be characterized by the number of "minconvex" functions applied to exhaust a concavity.

# 7. Dales

7.1. All operations in T to construct DF look symmetrical, they have 4 or 8 symmetrical parts in the formulas. What happens if only one part is used. For example, "**expand_1234_1236**":

d1234 = min (e1,e2,e3,e4);
d1236 = min (e1,e2,e3,e6);
if ( e5<d1234 || e5<d1236 )
  e5 = max(d1234,d1236);

7.2.1. Original connected set:                                          7.2.2. Result of the operation "expand_1234_1236":

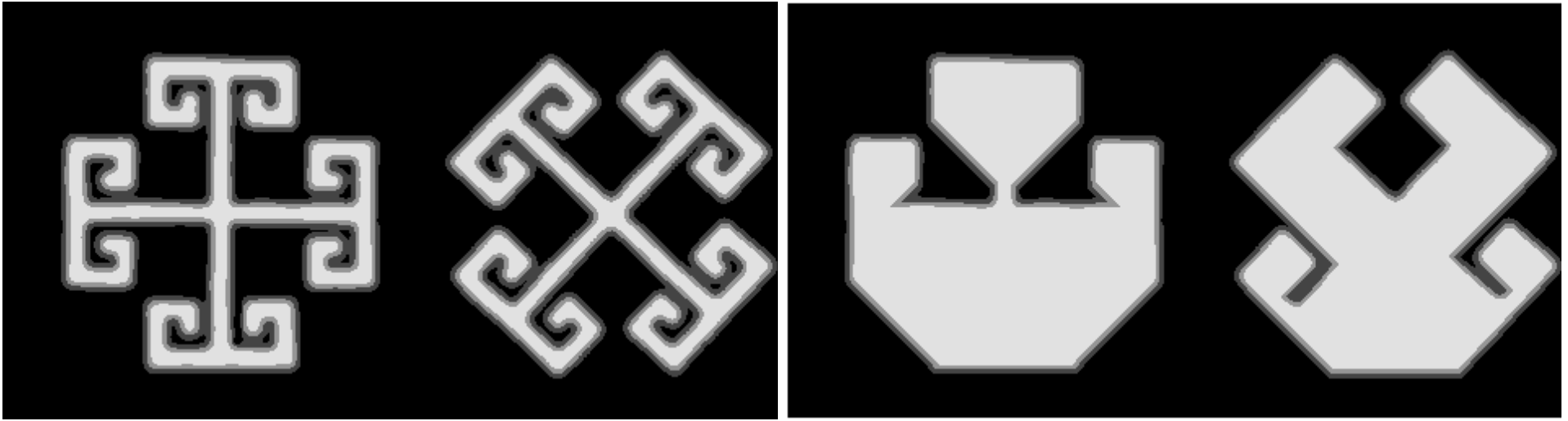

7.2.3. Difference

7.2.4 XOR

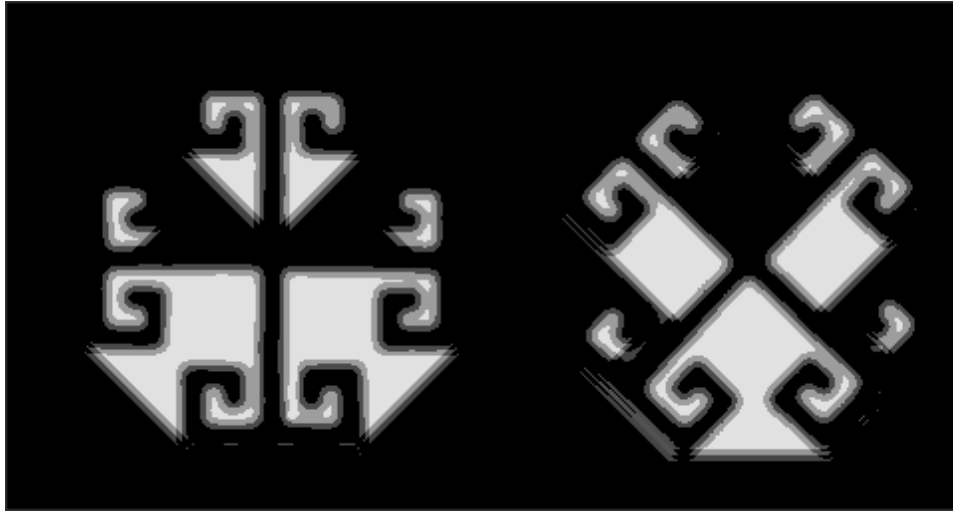
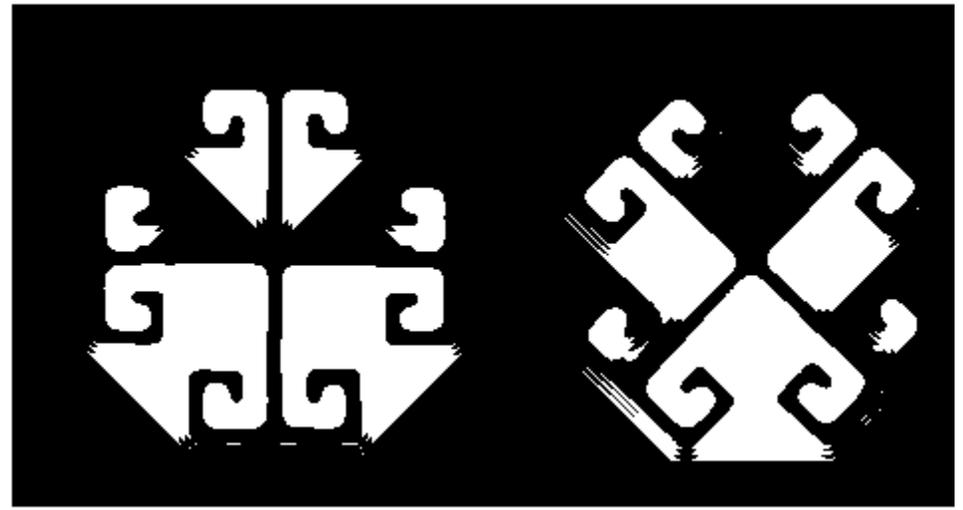

7.3. I'd like to call the components, which are complements of the sets obtained as the result of such "expand" operations, as "**dales**". Dales are the components which are DF among the tiles which are arranged in a certain way.

**The difference between Concavities and Dales is this.** While the construction of concavities require a convex hull be constructed before, dales are the intermediate stages of construction of the convex hulls themselves. Concavities are DFs built on the borders of a convex hull within the constraints defined by the original set, while the dales are DFs built on the ordinal set. Concavities "are dug", dales - grow.

It is convenient to call the dales by the direction to which they are "open". For example the **expand_1234_1236** can be called "filling the dales opened down". Because the set "grows" mostly in the down direction. I prefer to use "up", "down", "left", "right" names for directions instead of North, South, East, and West since the algorithms will be applied to classify symbols, see later, and "up and down" words seem more suitable.

Dales as connected components can be separated with a simple Difference operation.

There might be several kinds of dales, 4- or 8-connected. There are more dale examples further.

7.4. As with convex hulls, the dale "expand" operations can have their complements: "clean" which work on two tilings.

Here is one "clean" operation "**_clean_48_68**"
```
if ( e5 > t 5)
   d48 = max (e4,e8);
   d68 = max (e6,e8);
   if ( e5>d48 || e5>d68 )
```

d5=min (d48,d68);
applied after the expand_1234_1236 ( see 7.1)

7.4.2. Original two connected sets:

7.4.3. Result of the operation "expand_1234_1236":

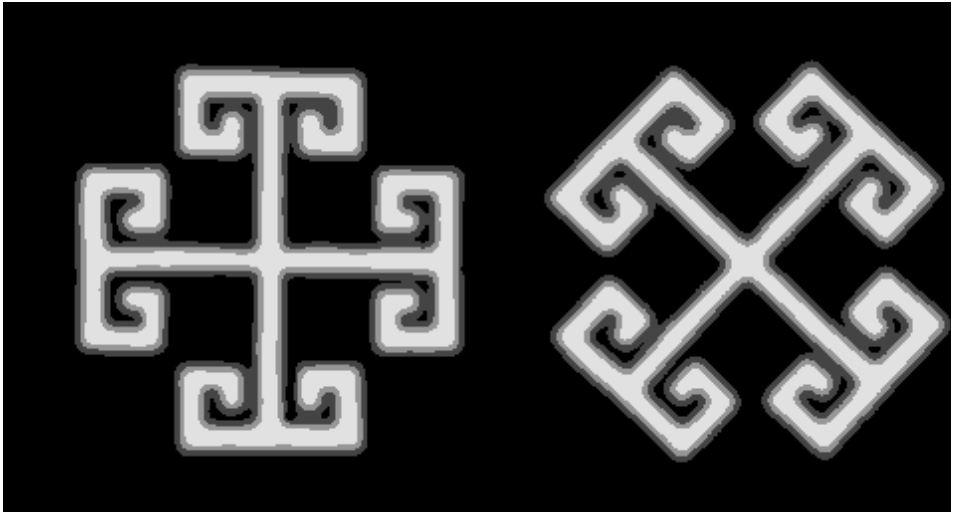

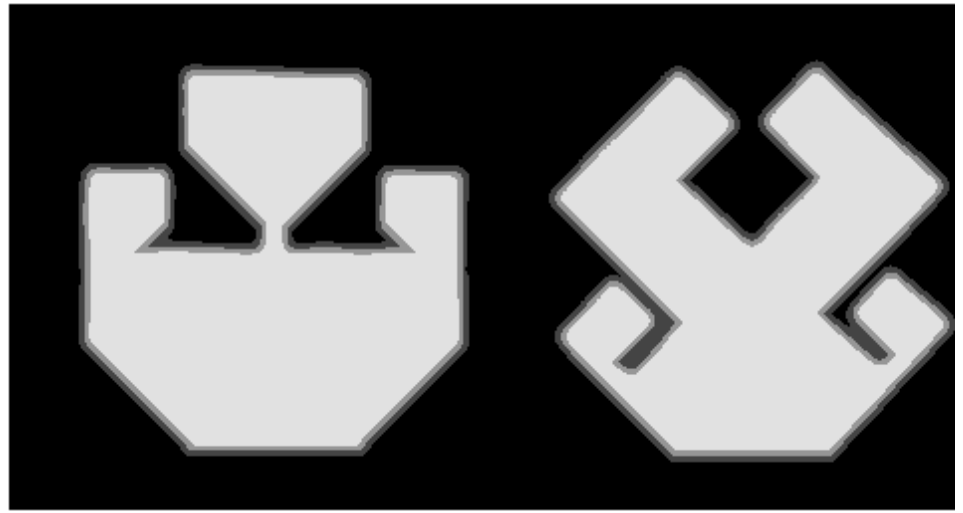

7.4.4. Result of the operation "**_clean_48_68**" applied to 7.4.3:

7.4.5. Difference with 7.4.2

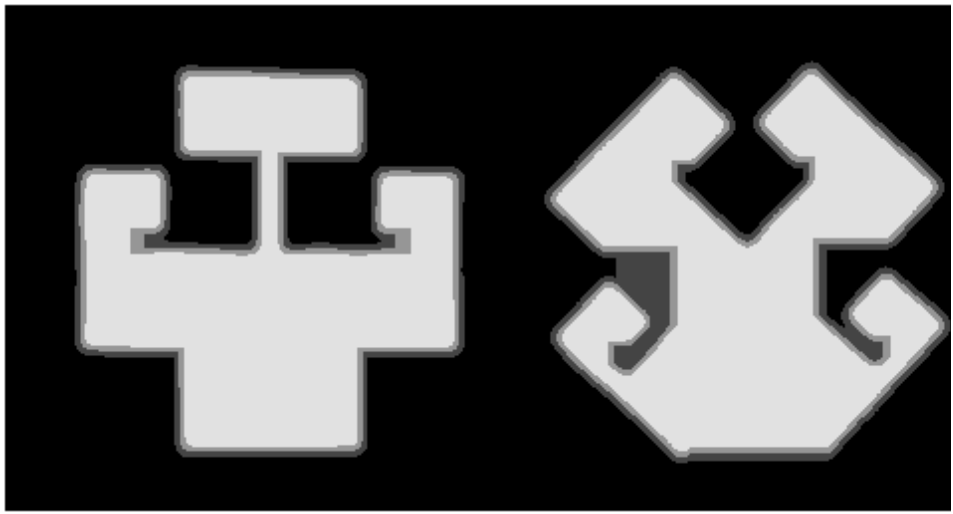

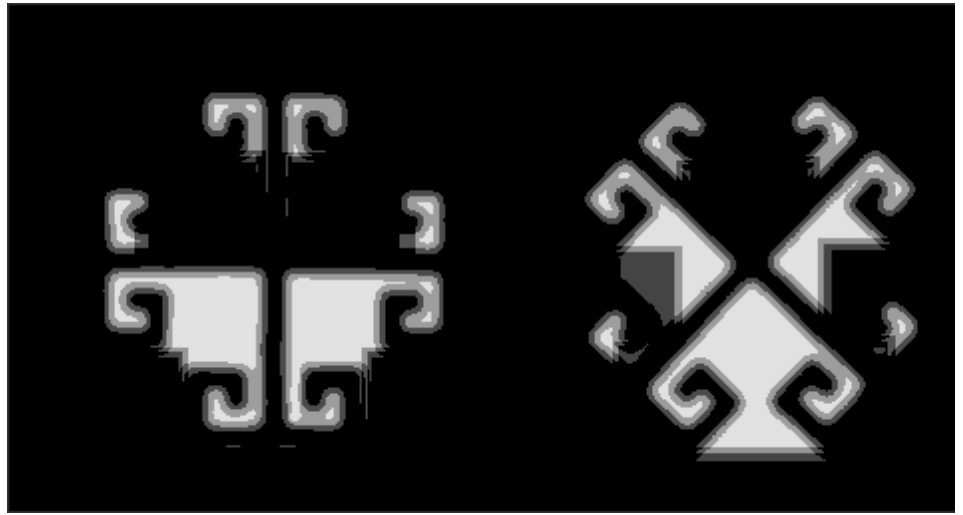

One can interpret the result as cleaner, more compact (in the intuitive way), smaller dales.

# 8. Directions, Rays, Angles in Quadrilateral Tiling Space

8.1. The relation of direction comes from the nature of the quadrilateral tiling T. The neighborhood of a tile define 8 major directions. Let us choose one neighbor, say e2. Then the following function "draws **a ray**" from any e5>0 to the "down" direction:

if (e5<e2) then e5=e2;

8.2. Other directions can be implemented in T as well. For example, let B==0 and for one tile e, V(e)==5. Then, the function:
for any e, if (e4)>1 then V(e)=V(e4)-1 or if (e8)==1, then V(e)=5;
"Draws a ray ", with tg(α)==1/5.
The method of variable values illustrated here is interesting and can be used in many ways. For now, let us confine ourselves to the major directions.

8.3. The function if (e5<e2 || e5<e1) then e5=max(e2,e1) fills up "**an angle**". The affected area is spreading in the directions opposite to e2 and e1.

8.4. The similar function may be made for up to 5 consecutive neighbors, like e4,e1,e2,e3,e6. In this case the resulting set takes half of the tiling space.
If the 6-th neighbor is used, the whole tiling space will be filled. I believe, this is an important distinction, five neighbors or six, producing "Half space" vs "Whole space" which can be used for the further research.

8.5. Here are the results of the mentioned operations:

8.5.1. Original 256-level image:

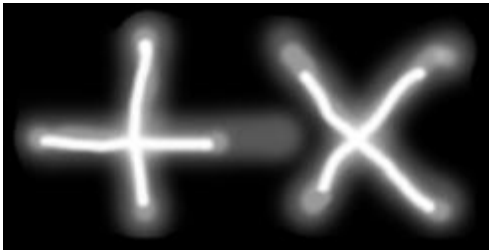

8.5.2. The result of the 8.1, the set formed by "rays":

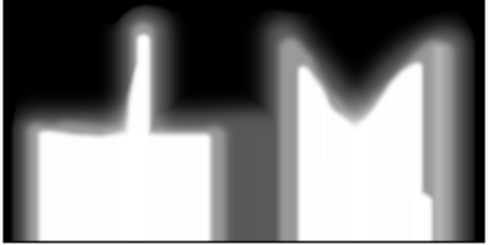

8.5.3. The difference:

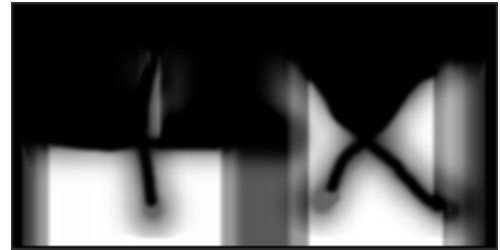

8.5.4. The result of the 8.3, set formed by "angles":

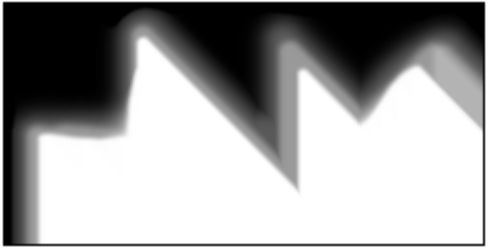

8.5.5. The difference:

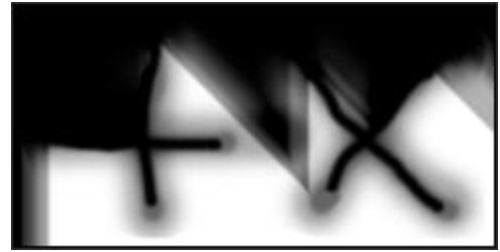

8.6. "Ray" operations can be used to decompose the space into regions (colors) so that mutual relations among different regions can be found, like "above", "on the side" and so forth. There are many of such relations, and there many more if 8.7 is taken in account.

8.7. Constraint operations with rays and angles can be of different kinds.

8.7.1 Here is the Constraint Ray-type operation **_ray2_2**() which "grows rays out of dales with the original image as a constraint. This operation may be useful to distinguish between different types of dales.

```
if ( e5 >= t5 )
   if (e5 < e2 )
     return e2;
```

8.7.1.1. Result of 8.7.1. The T1 is the cleaned convex hull 7.4.4, T2 is the original 7.4.2.

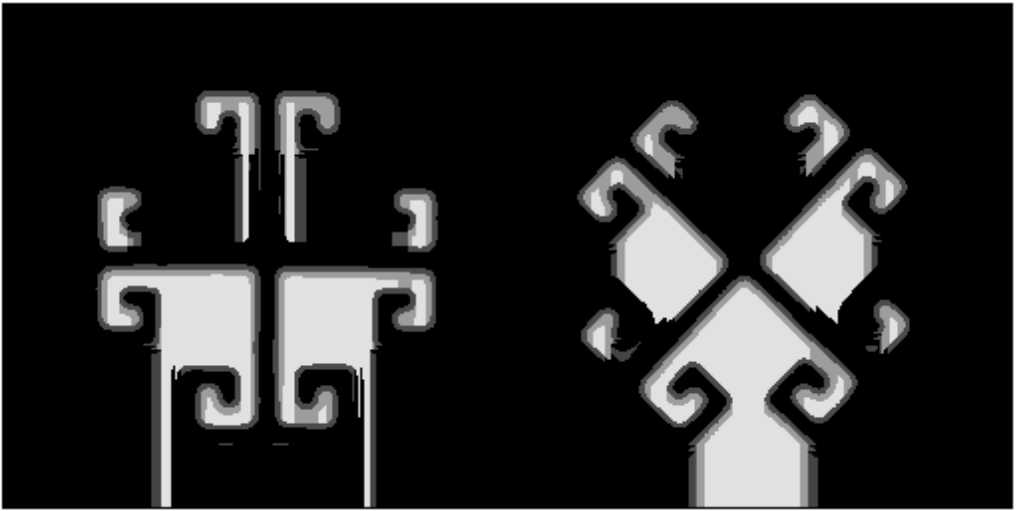

8.7.2. Here is the operation **_rayanti_8**() which "cuts the rays out of connected components with the original image as a constraint". This operation is useful in investigation of a dale structure.

```
if ( e5 > t5)
  if (e5 > e8 )
    return e8;
```

8.7.2.1. Result of 8.7.2.  
The T1 is the cleaned convex hull 7.4.4, T2 is the original 7.4.2.

8.7.2.2. Difference with 7.4.2

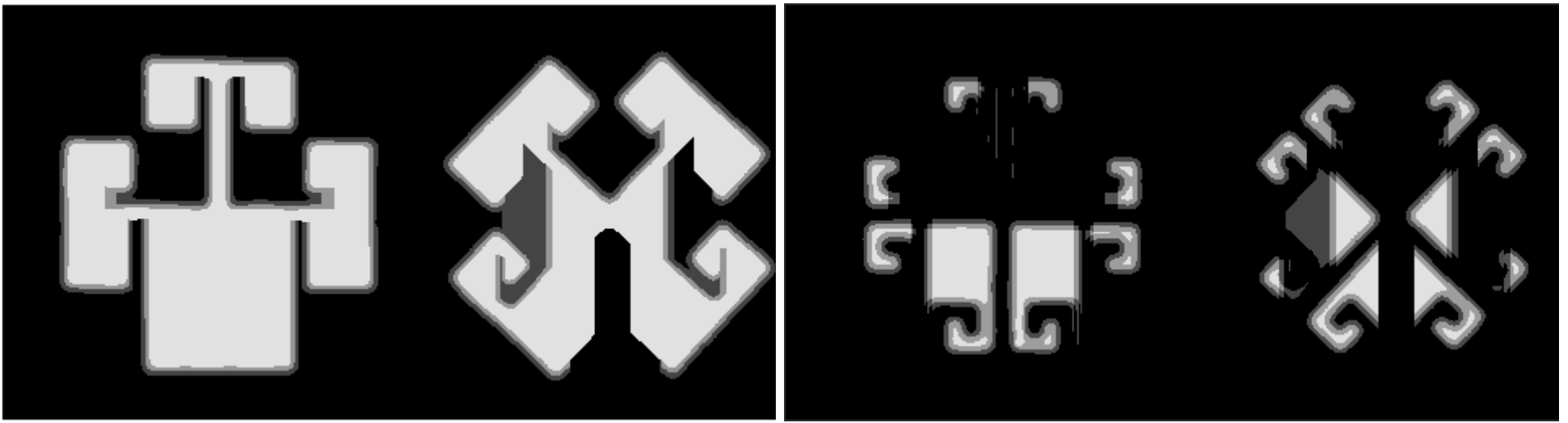

## 9. Component and Relation Matrix Instead of Concavity Tree

Concavities, dales, rays, angles as independent connected components can be obtained by finding the differences between results of various operations. These differences form a wide variety of relations. They include such relations which intuitively can be described as "above", "within". It is possible to find components which are "**lakes**", and "**dales of dales**" and apply the same operations to them and so forth. Overall, it is much more than a "concavity tree" [2]. There is a reason to talk about "component and relation matrix".

## 10. Hypothesis About the Nature of Image Analysis in Living Organisms

10.1. Here is the hypothesis: "Living organisms use dales and their mutual location to recognize images. The symbols we use, in any culture, consist of dale combinations. The strokes are used to draw symbols, but the symbols are recognized not by the strokes, but by dales they form. The importance of the eye tests like **Landolt C** is well known in ophthalmology."

10.2. I'd like to point your attention to the high robustness of the dale features: they are independent of size, general orientation, structure, texture, font of symbols, unevenness of gray levels, other distortions, and noise. Here is the table from [1] which lists the most important features of the latin alphabet. A "dale opened down" is a connected component shown at 7.4.5.

| Symbol, | Features |
|---|---|
| A | A lake over a dale opened down (dale-down) |
| B | Two lakes above each other. One small dale-right on the right |
| C | One dale-right |
| D | One lake. A dale-left on the left (compare with O). See 14. |
| E | 2 dales-right |
| F | A dale-right over a dale-down-and-right |
| G | A dale-right with a dale-right. (dale of a dale) |
| H | Dale-up over a dale-down |
| I | Dales left-up-down and right-up-down. See 11. |
| J | Dale-up and -left |
| K | Tree dales: up, right, down |
| L | A dale-right and up |
| M | Three dales: two dales down and one up in between |
| N | Two dales: one up, another down. They are to the sides of each other ( compare with H ) |
| O | One lake, no dales |
| P | One lake above a dale opened right and down |
| Q | One lake, small dales at the bottom-right |
| R | One lake, two dales below: one opened right, another down |
| S | A dale opened right over a dale opened left |
| T | Two dales to sides of each other, one to right and down, another to left and down. |
| U | One dale up. |
| V | One dale up. Two small dales below opened down and to sides (compare with U)( See O and D ) |
| W | Two dales up, one down in between. |
| X | Four dales... |
| Y | Three dales... |
| Z | Two dales...directions are contrary to S |

10.3. Tile space and operations on it presented in the paper are well suited to be used by the living organisms: the operations are performed on the network which is similar to the neuron networks of human neural system, they do not rely on mathematical operations except the most simple operations on magnitudes: max and min, difference, inversion. These operations do not require synchronization: they can be performed in the truly asynchronous way. The chance that these types of operations are used in the living brain feels higher than the chance of synchronous operations.

10.4. The tiling may be realized as a matrix of simple processing elements (PE) [1] as it is shown on the Figure below. The term "graph" on the diagram is a synonym of "tiling". I am in the process of defining the best terms. Let me point out, that this is one PE of an asynchronous parallel processor which may contain hundreds or millions of PEs.

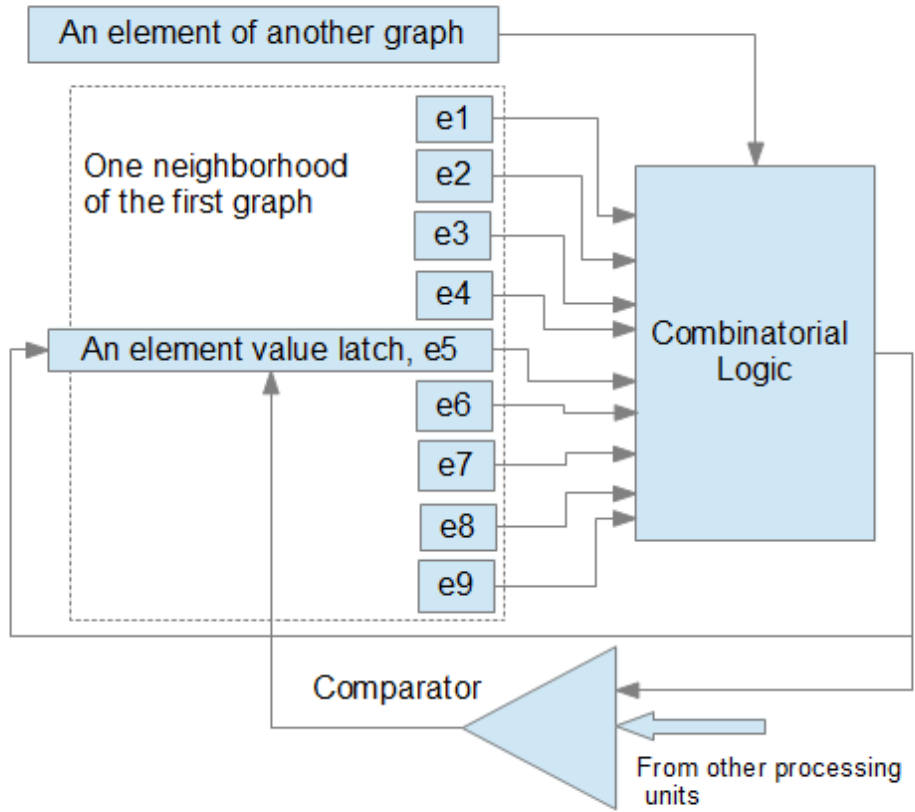

## 11. Ridges as Dales of Dale-Derived Components

11.1. If dales are called dales, than it is naturally to call the components which immediately contain dales as **ridges**. Those which may be called Stems, Strokes, Tails, Arms, Legs, or Bars. According to the hypothesis, they are the secondary elements and are found by their relation to the dales and lakes. There are several methods possible to find them. I am going to present two methods which illustrate well the propagation operations as examples.

11.2. Extracting the left stem of the letter "A".

11.2.2    11.2.3    11.2.4    11.2.5    11.2.6    11.2.7    11.2.8

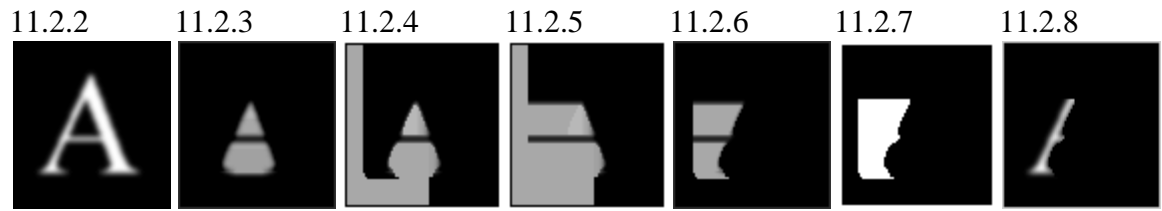

11.2.2 - The original gray image;
11.2.3 - the dale opened down of 11.2.2 as a separate component obtained with **_expand_1234_1236()** and then **_clean_46()**, and then minus(). Note that 11.2.3 is a single component since the bar of A is slightly darker than the white stems;
11.2.4 - the angle expansion of the 11.2.3 obtained with **_ray_268()** and 11.2.1 as T2;
11.2.5 - the filled dale opened up of 2.4 obtained with **_expand_4789_6789()** and then **_clean_46()**.
11.2.6 - 2.5 minus 2.4. This is the ridge;
11.2.7 - XOR of 11.2.5 and 11.2.4;
11.2.8 - if 11.2.2 contains a separate component (gray on the black background), a more precise stem can be found with a thresholding like **_thresh_value()**.

11.3. Vertical stroke: a ridge as a dale.

11.3.2    11.3.3    11.3.4    11.3.5

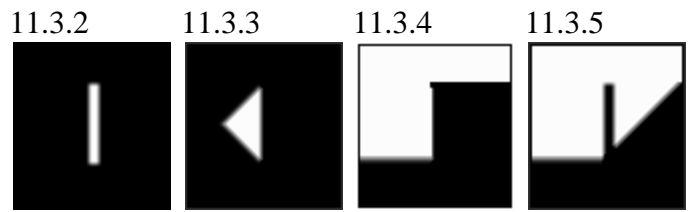

11.3.2 - The original gray image;
11.3.3 - the dale to the left obtained with **_expand_369()**;
11.3.4 - the angle expansion of the 11.3.3 obtained with **_ray_468()** with 11.3.2 as T2;
11.3.5 - the dale to the right of the 11.3.4 obtained with **_expand_147()**, then minus 11.3.2. The final 11.3.5 image contains a dale opened down which can be found.

11.4. **_ray_268()** propagation operation is

```
if (e5>=t5)
   if (e5<e2 || e5<e6 || e5<e8)
      e5 = max(e2,e6,e8);
```

11.5. **_ray_468()** propagation operation is similar to _ray_268()

11.6. Concluding, the ridges are the dales of components obtained as the result of

- Dale operations ( Convex Hull - type )
- Clean operations ( Minimized Convex Hull - type )
- Ray operations

# 12. Waterfall. Operations of Allowance. Difference with operations of Constraint

12.1. The difference between operations of constraint and allowance is the following:
Constraint operations change an image T1 containing information. Changed areas of the image differ from the corresponding areas of the reference image T2.
Simple Allowance operations fill up "an empty image". Filled areas of the image are equal to the corresponding areas of the reference image.
More complex Allowance operations may fill up an image separated into areas where the operation is allowed. In the simplest case, it is tiles with value 0, i.e. "an empty image". The operation _ray_268() presented above is an operation of allowance.
Obviously, for an operation to propagate, the areas of allowance should be connected. The operation will propagate until there is no more allowed areas or the reference components are exhausted.
There could be many operations of Allowance, for example 4- or 8-connected, in a certain direction, upon different allowance areas, and with different starting components.

12.2. As an example, I'll present an operation of Waterfall. T1 is an image to propagate in, T2 is a reference image (t5 element belongs to T2). This particular function is called **_waterfall_min()**

```
if (e5 is allowed)
   e = max (e2, e4, e6, e8);
   if ( t5<= e)
      e5=t5; disallow e5;
```

This Waterfall fills up the area in the image T1 corresponding to the certain monotonic area in the reference image T2.

12.3 Here is an example. Waterfall was used to remove the monotonic areas adjacent to the border of the image, obviously, they do not convey useful or undistorted information.

12.3.1. Original Linda.png image smoothed with Gaussian r=5.0, T2    12.3.2. Result of waterfall_min(), T1    12.3.3. Difference

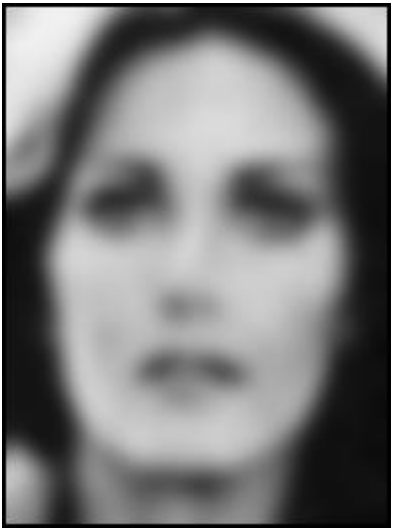
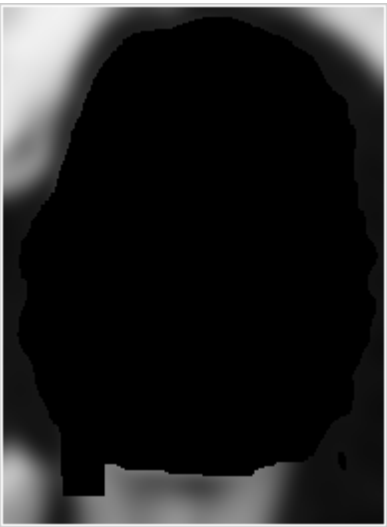
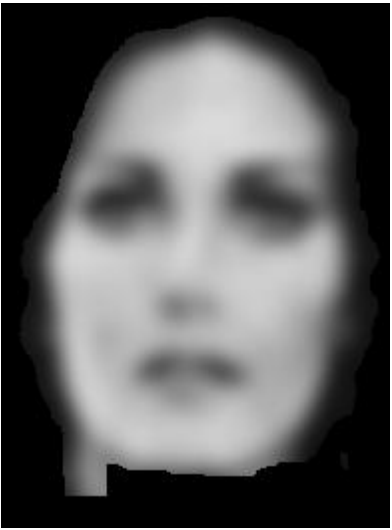

## 13. A Tip Phenomena in the Leveled Tiling Space

I'd like to point your attention to the tip phenomena which is especially profound on leveled images. Here is the 4.16.4 again:

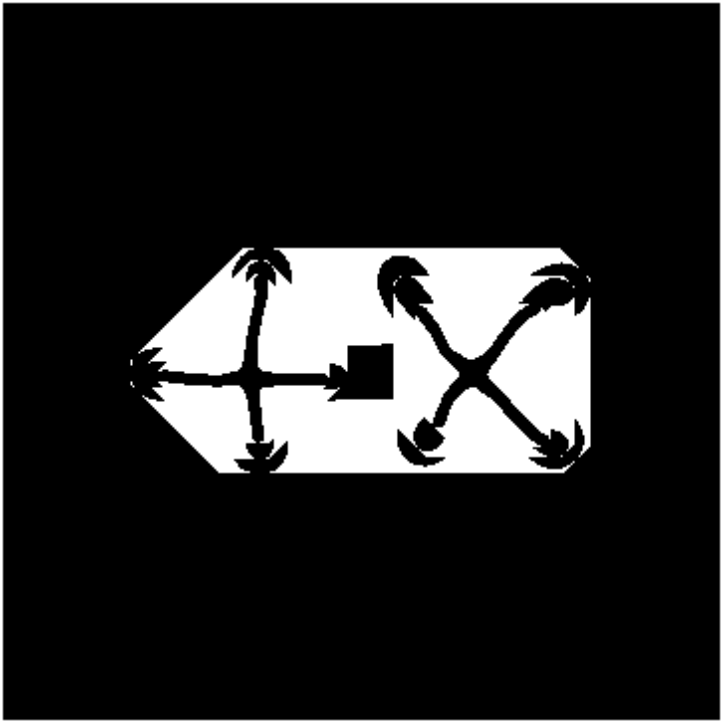

## 14. DO Example. Dales of Sides

14.1. Here is an example of the operation "**Expand 369**", the rule is

for ∀e5, if ( e5<min(e3,e6,e9) ) ⇒ e5 = min(e3,e6,e9)
repeat;

and the operation "Expand 689", the rule is

for ∀e5, if ( e5<min(e6,e8,e9) ) ⇒ e5 = min(e6,e8,e9)
repeat;

14.1.1. Original connected set (8 levels):   14.1.2. Result of the "Expand 367":   14.1.3. Result of the "Expand 689 ":

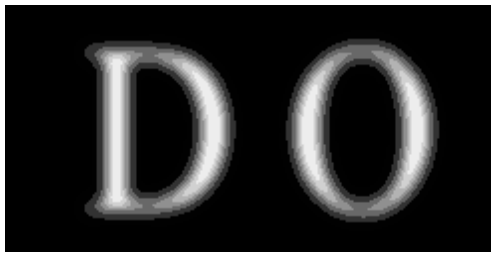 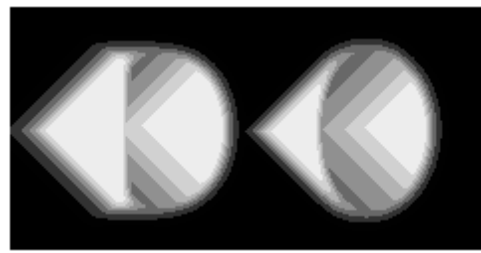 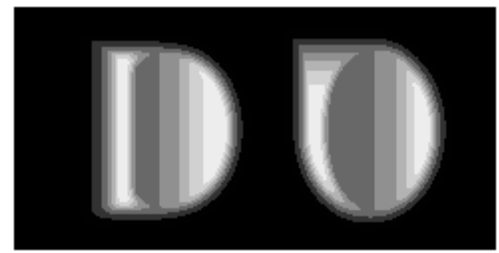

Notes. The original example image has 8 levels of brightness. It contains the printed black-and-white "Times New Roman" font characters which were smoothed by averaging and then levelized. Note 14.1.2: in addition, dales opened up and down can be filled on it. It may give some interesting results: the difference between images may be more vivid. The thickening of the vertical strokes is widely used in typography.

14.2. Dales can be found on the **differential images**. It is essentially the shape analysis of the certain "sides" of components, see the example below. The differential function **diff** used is:

if (e5<e6)
   e5_2 = e6-e5;
else
   e5_2 = minimal_value;

This function is not a propagation function, the result of it is stored in another tiling. The diff result shown on the 14.2.2 was amplified for better visual presentation.

Then the "dale opened right" function was applied. The function has a slightly different form, which is easy to grasp from its name (see 2.3). This one can be used on "x"-connected components.

14.2.1. Original connected set (256 levels):

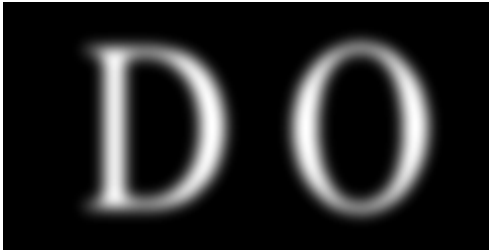

| 14.2.2. | 14.2.3. | 14.2.4. |
| Diff of the 14.2.1 in the e8 direction: | Result of the "_expand_247_148": | Difference: |

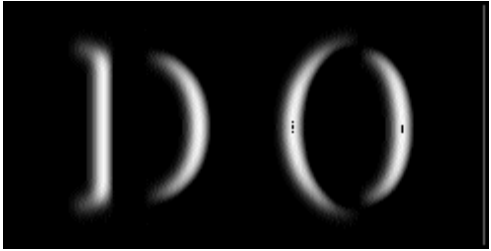 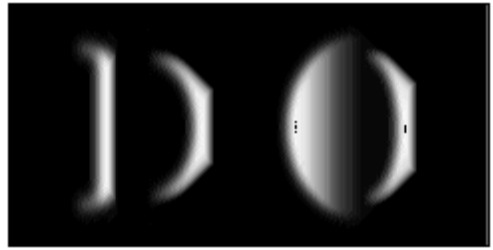 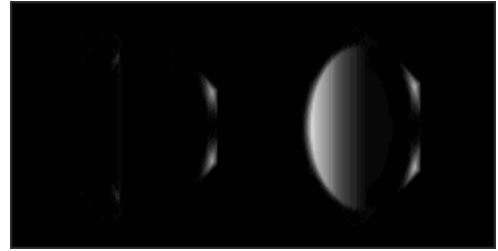

## 15. An Example with the Latin Alphabet

Limited by space and time, I can not present the complete algorithms of partitioning the set of latin letters into to the corresponding set of tile features. Here is an example of the most general operations. Note that all operations were performed on the whole image at once.

15.1 The original alphabet image. It is typed in Photoshop(TM) and "Blurred more".

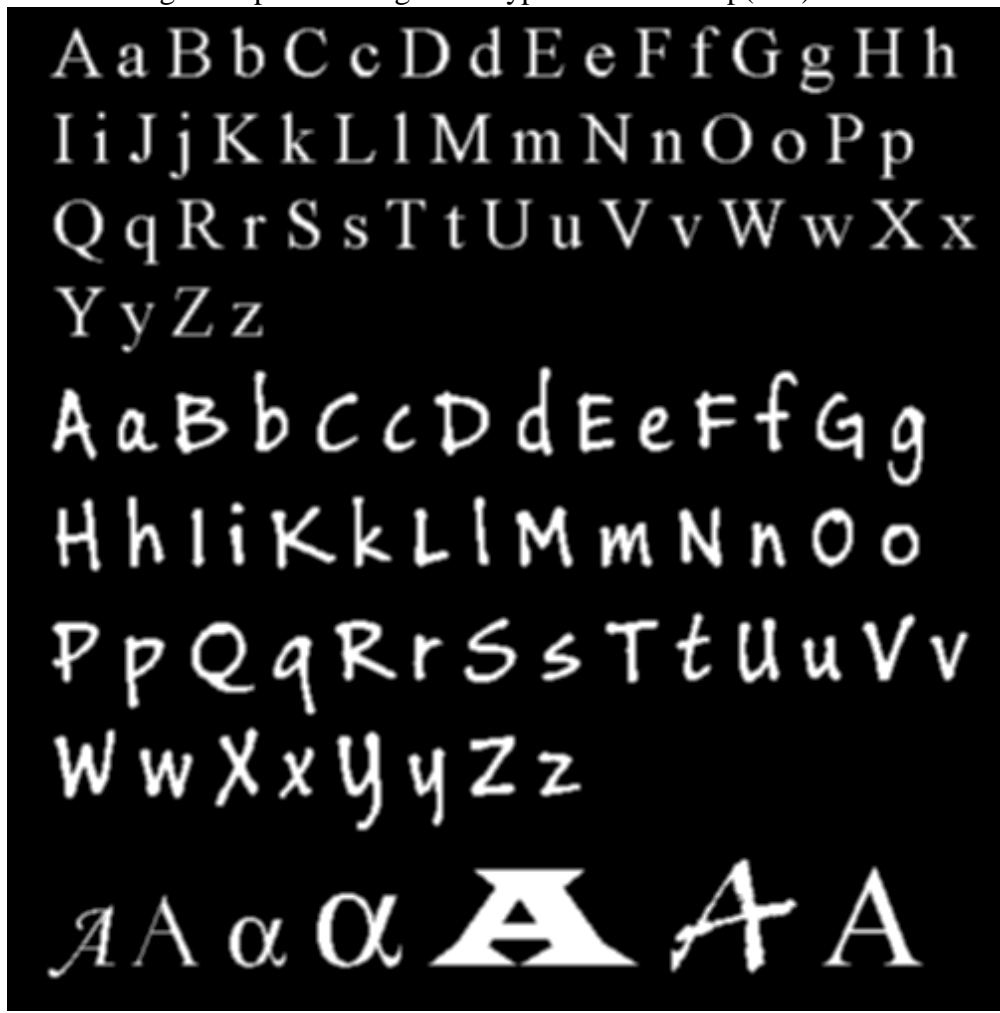

15.2 convex_hullB() on 15.1

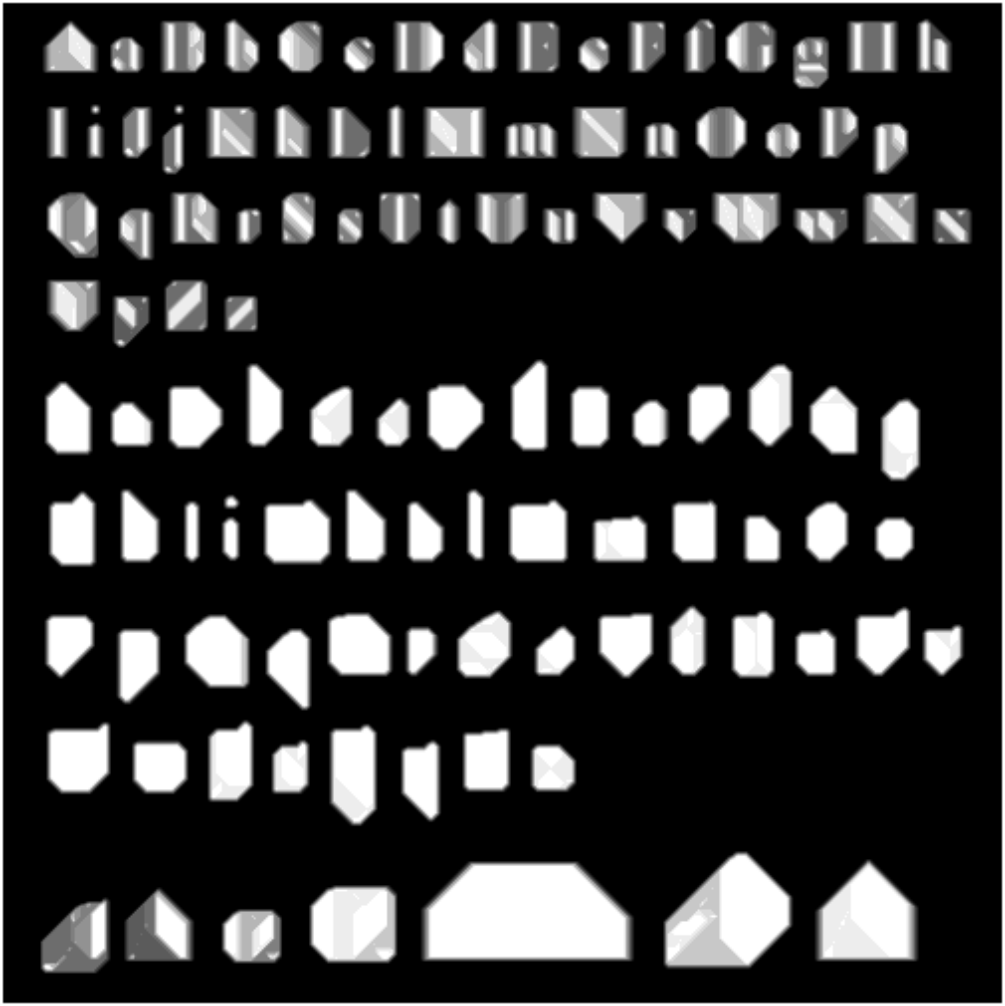

15.3 clean_123456789() on 15.2. This is the image of "lakes" filled.

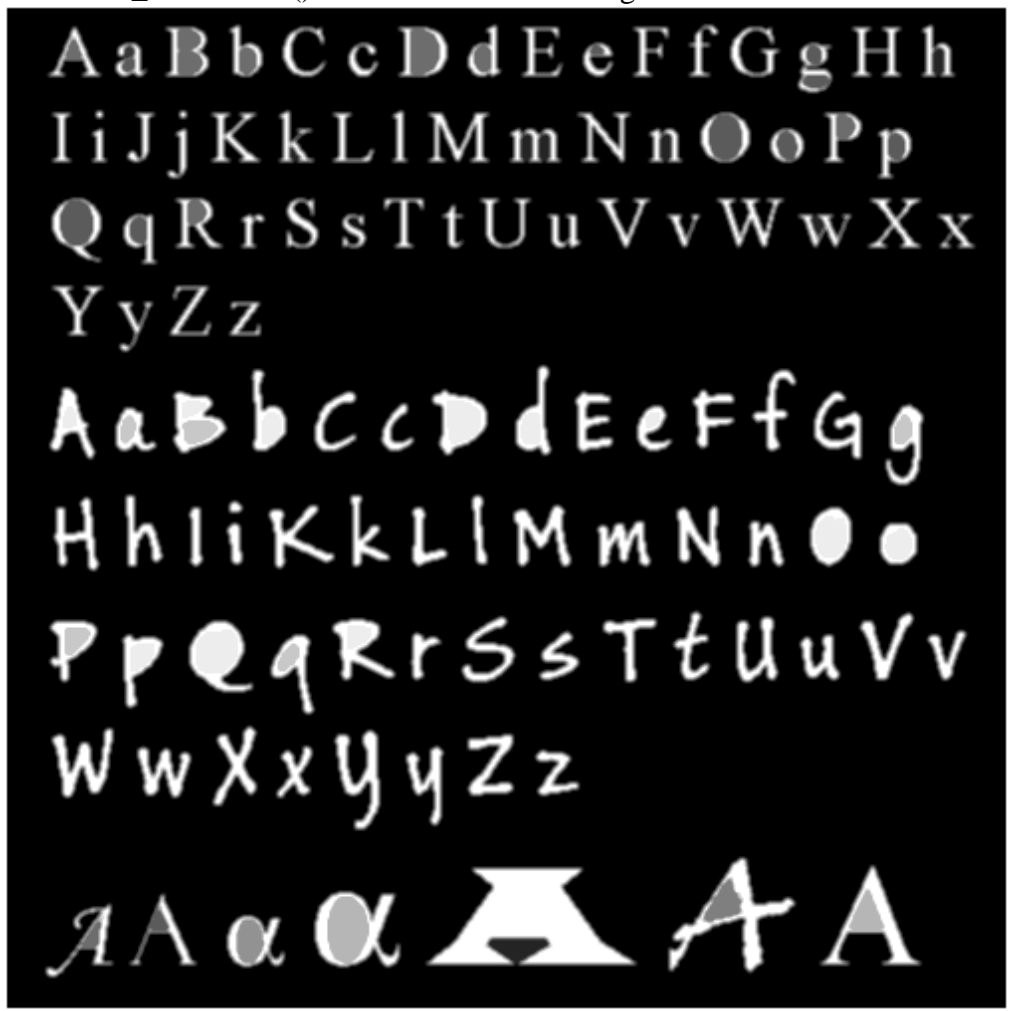

15.4. expand_1234_1236() on 15.1. This is the image of dales opened down and lakes.

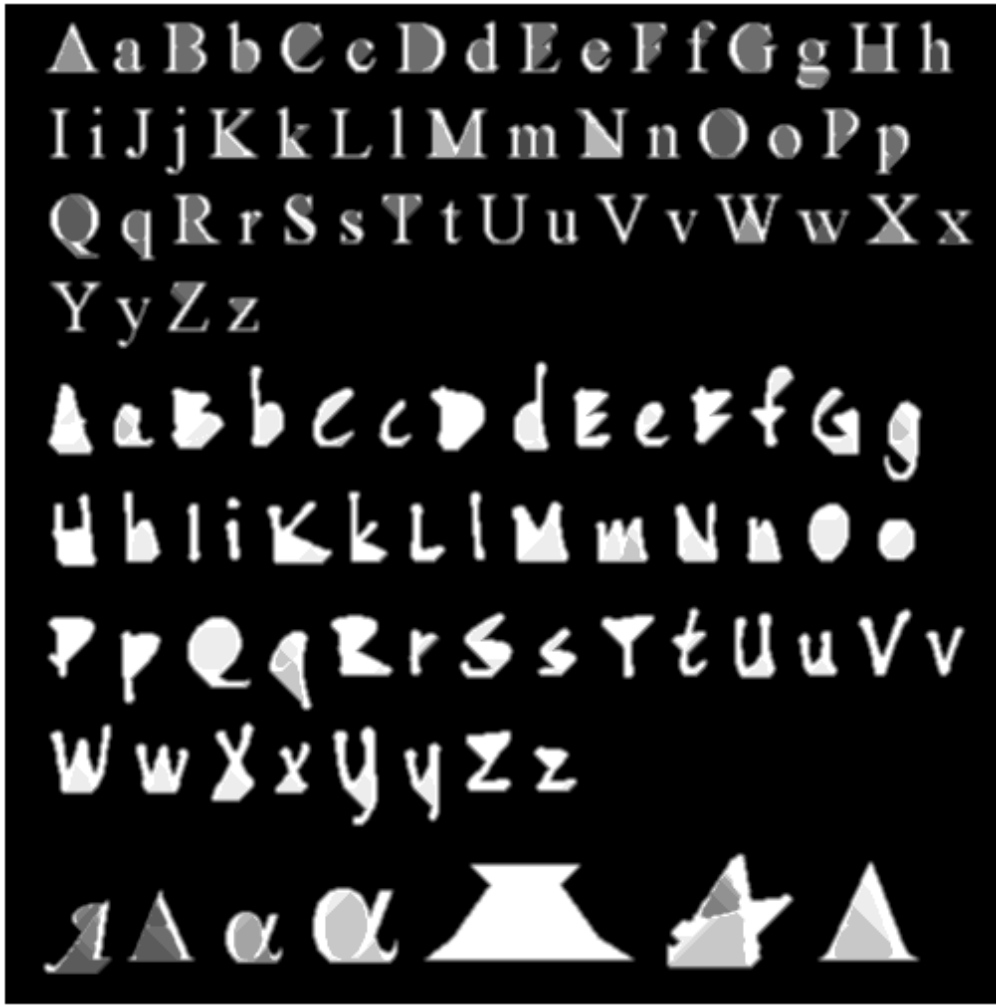

15.5. clean_46() on 15.4. This is the "cleaned" dales opened down.

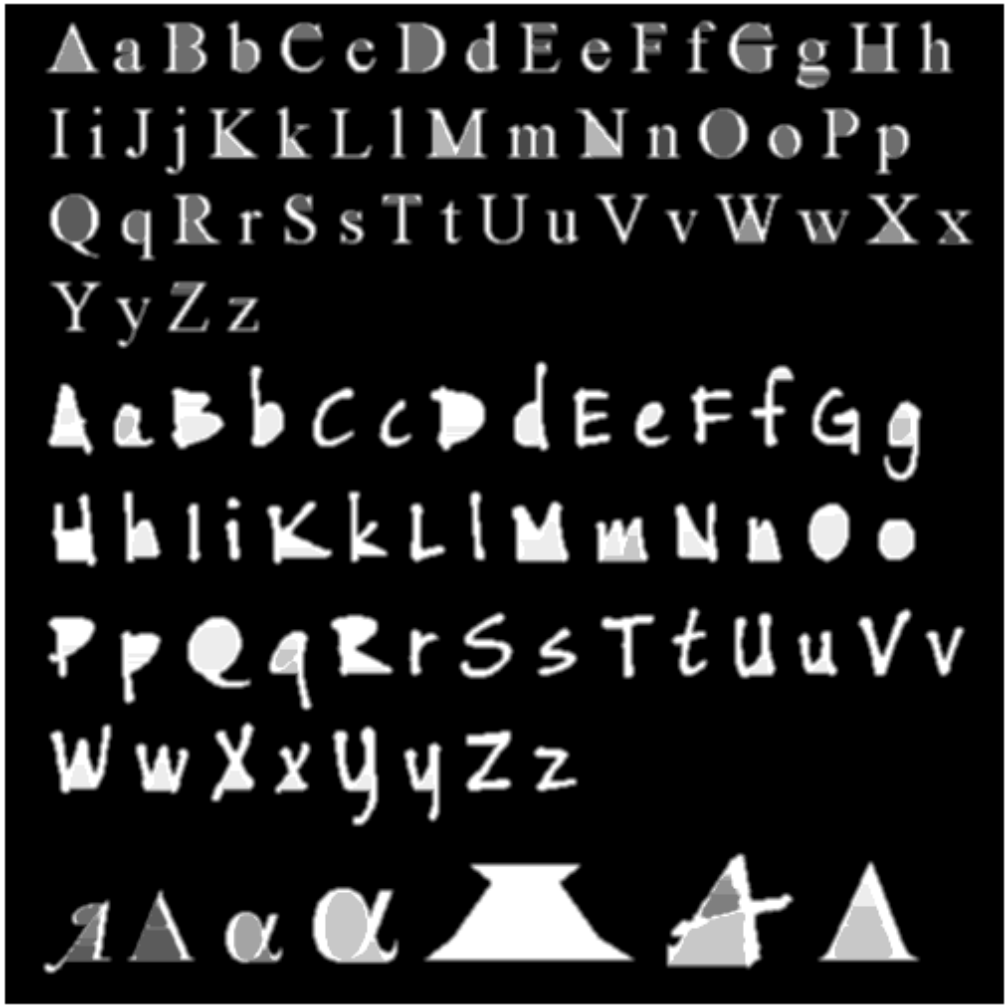

15.6. expand_1247_1478() on 15.1 and then _clean_28(). This is the image of dales opened right and lakes.

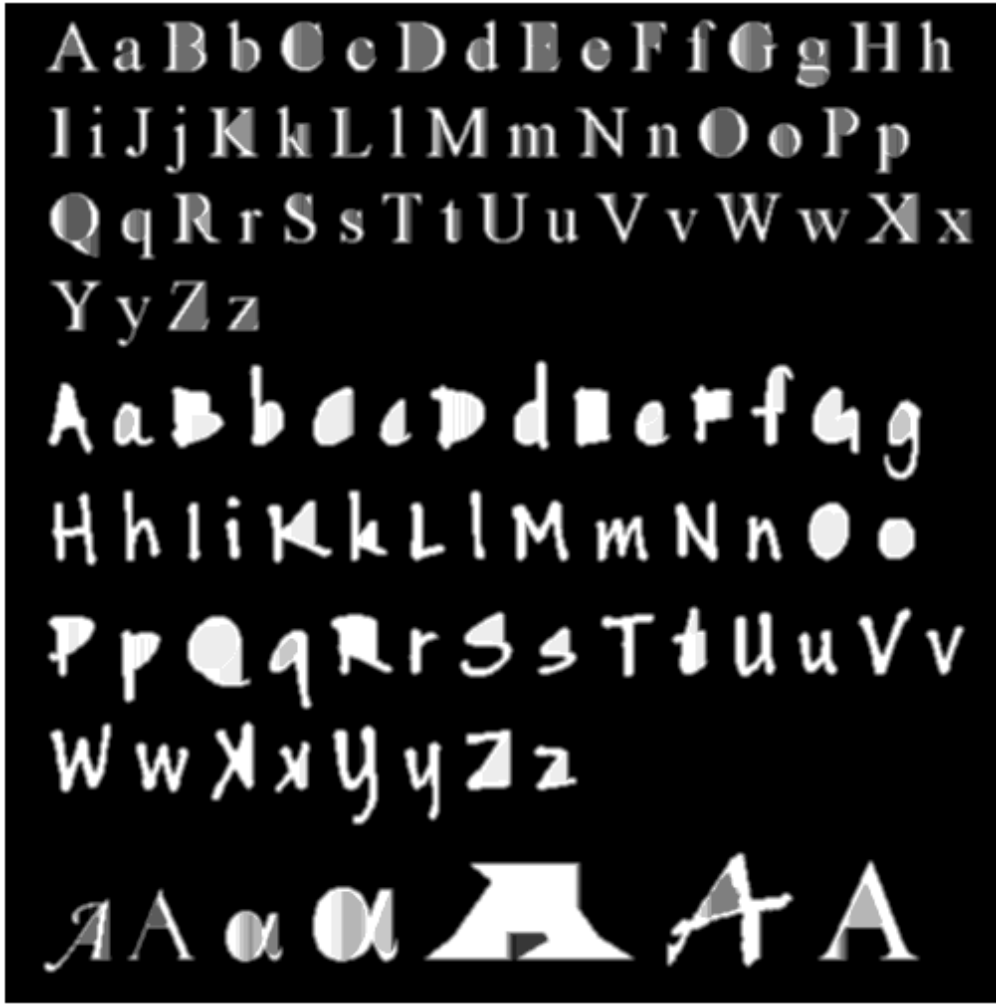

15.7. expand_2369_3689() on 15.1 and then clean_28(). This is the image of dales opened left and lakes.

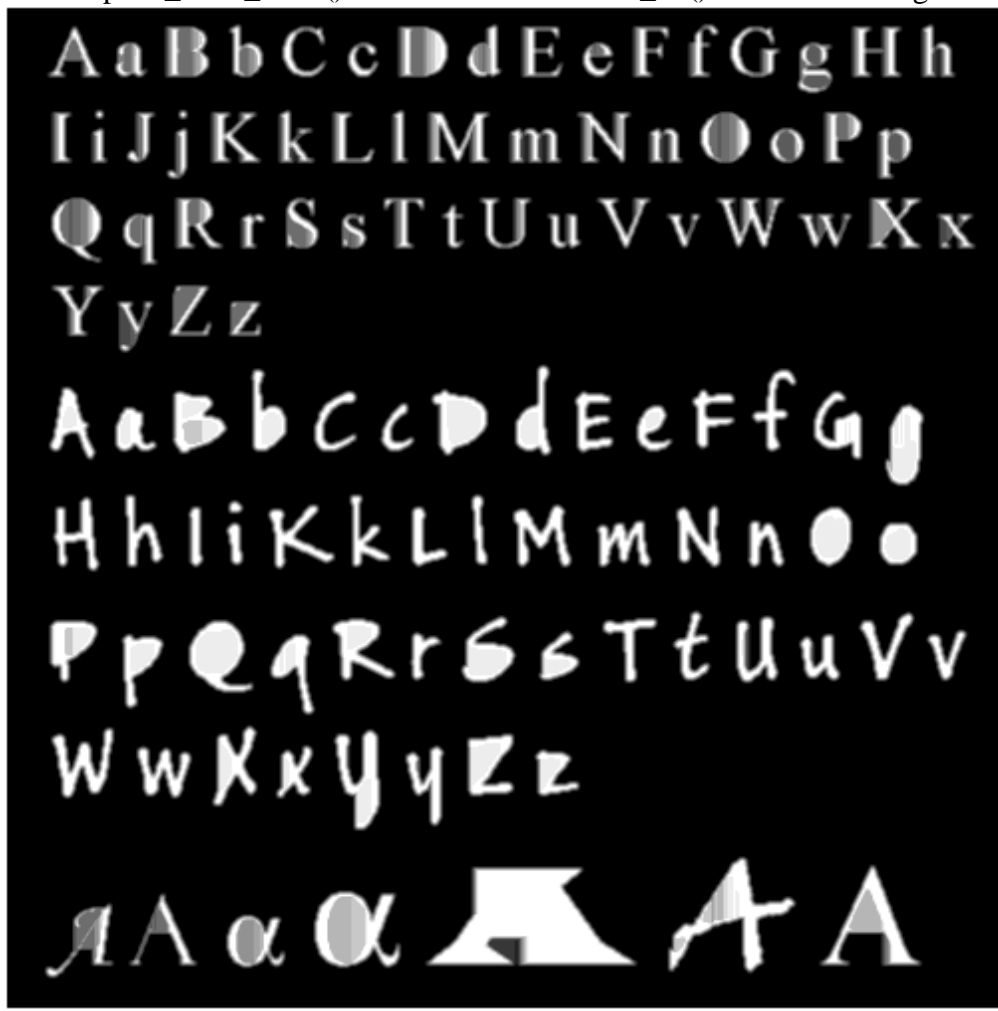

# 16. Hieroglyphic Example. Partitioning of the Gray Image with Dales and Lakes

16.1. The example images are self explanatory. Note that 16.2.1 is a gray image, no components are partitioned. Other images starting from 16.2.4 are all partitioned (because they originate from 16.2.4 which is the difference) into separate gray components on the 0-level background. The components can be counted and analyzed separately.
16.2.1 - the original image;

16.2.2 - convex_hullB() of 16.2.1;
16.2.3 - minconvex_hull() of 16.2.2;
16.2.4 - 16.2.3 minus 16.2.1;
16.2.5 - convex_hull() and minconvex_hull() of 16.2.4;
16.2.6 - _clean_12346789() of 16.2.5 with T2==16.2.4;
16.2.7 - 16.2.6 minus 16.2.4. It is the image of "lakes" of 16.2.4;
16.2.8 - dales opened down of 16.2.1 obtained with _expand_1234_1236();
16.2.9 - dales opened right of 16.2.1 obtained with _expand_1247_1478();
16.2.10 - dales opened left of 16.2.1 obtained with _expand_2369_3689();
16.2.11 - dales opened up of 16.2.1 obtained with _expand_4789_6789();

16.2.1
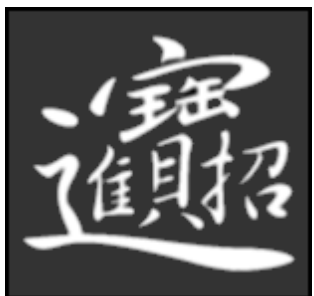

16.2.2
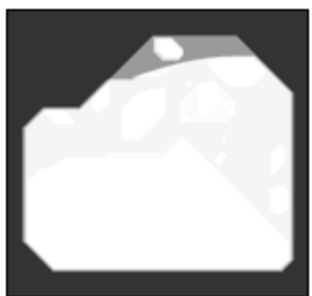

16.2.3
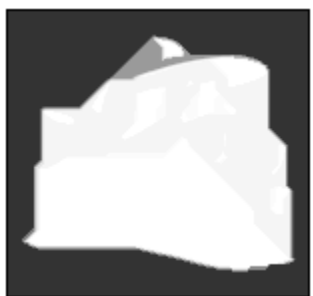

16.2.4
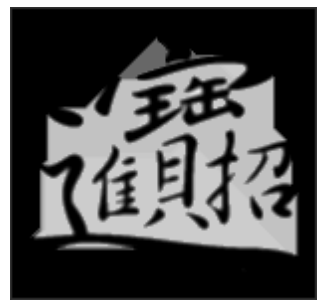

16.2.5
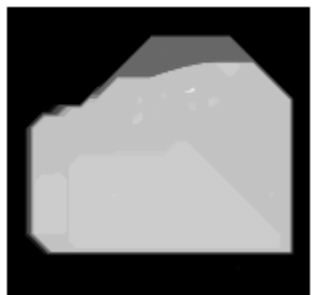

16.2.6
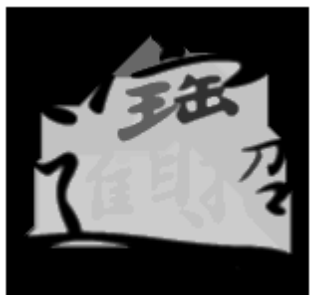

16.2.7
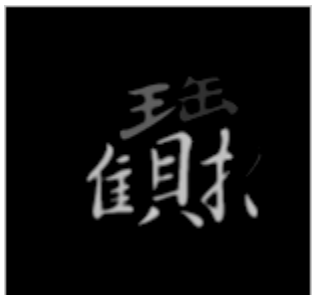

16.2.8
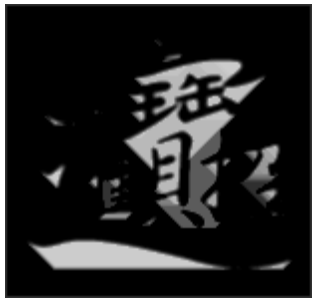

16.2.9
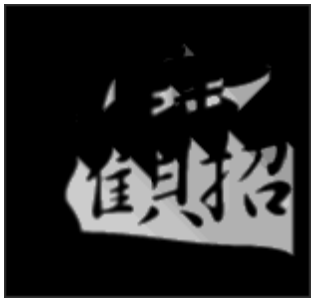

16.2.10
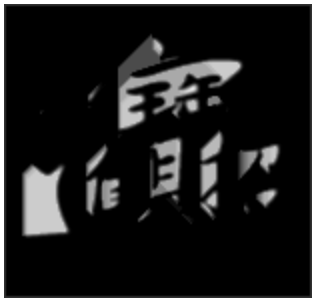

16.2.11
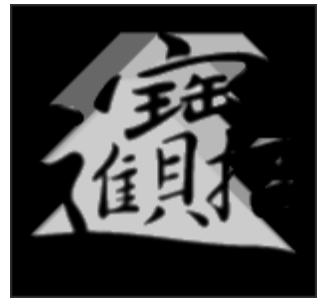

# 17. An Example on a Face

17.0. For the sake of curiosity, I'd like to present the results of dale operations on a human face. This shows application of the simple functions and it is very far from a comprehensive analysis. Essentially, it is only an illustration of the functions results on a more complex gray image.

17.2.1 Original Linda image T2 smoothed with Gaussian r=5.0      17.2.2. waterfall_min(), T1      17.2.3. 17.2.1 minus 17.2.2

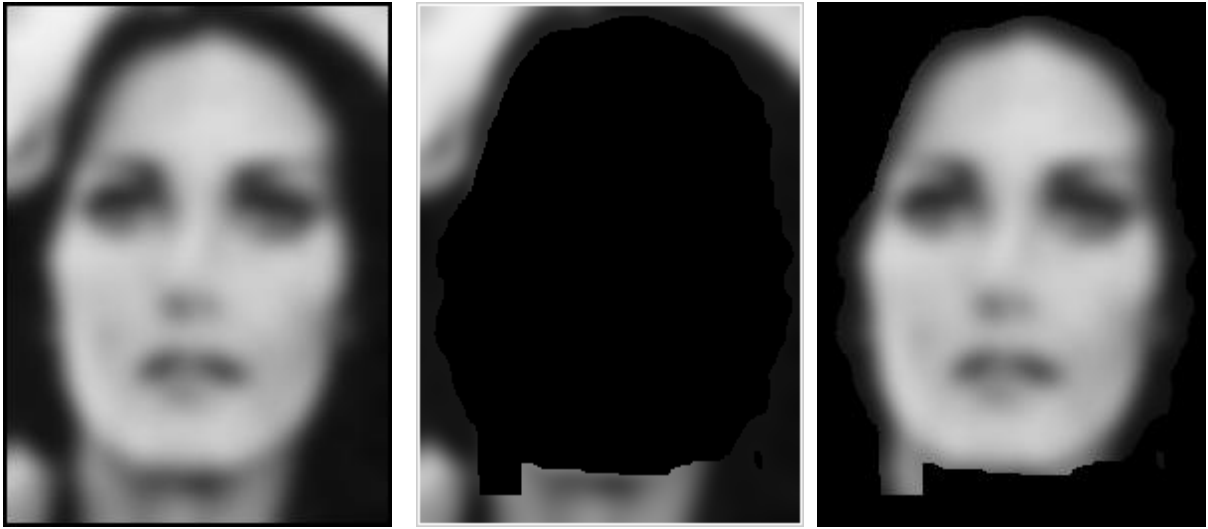

| 17.2.4. Convex Hull _convex_hullB() | 17.2.5. 17.2.4 after _clean_12346789() | 17.2.6. 17.2.5 minus 17.2.4 amplified | 17.2.7 XOR of 17.2.5 and 17.2.4 |
|---|---|---|---|
| 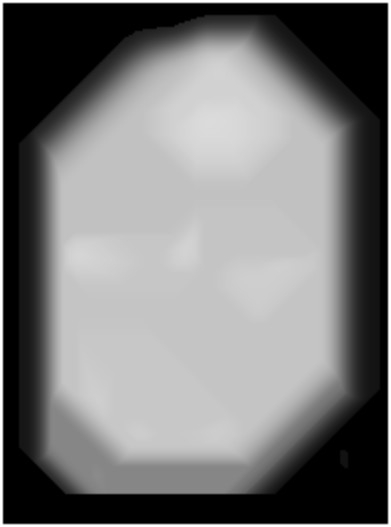 | 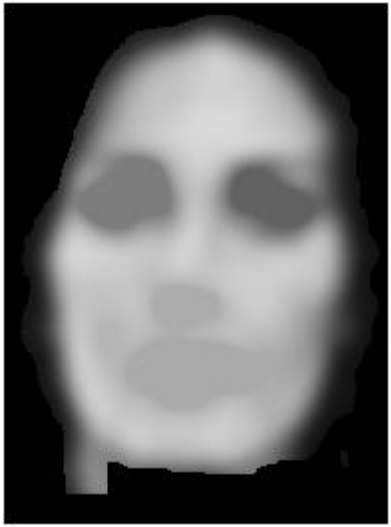 | 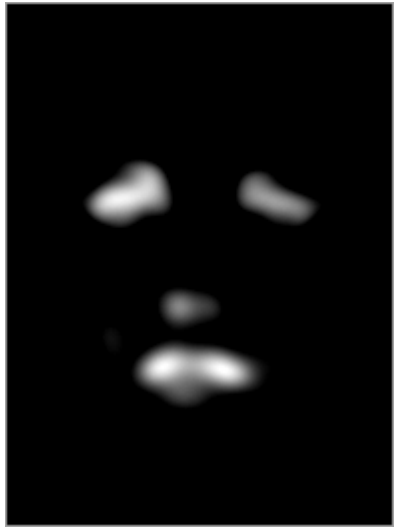 | 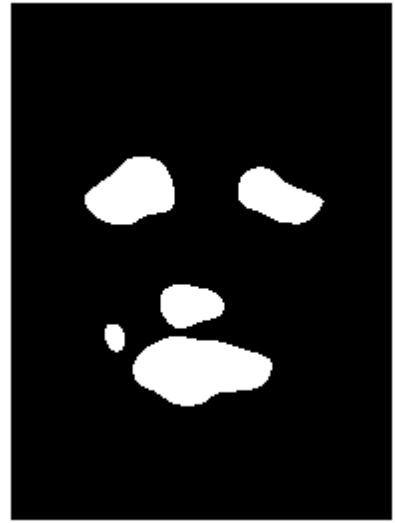 |
| 17.2.8. Convex Hull of 17.2.6 _convex_hullB() amplified | | 17.2.9. 17.2.8 minus 17.2.6 amplified | 17.2.10. XOR of 17.2.6 and 17.2.8 |
| 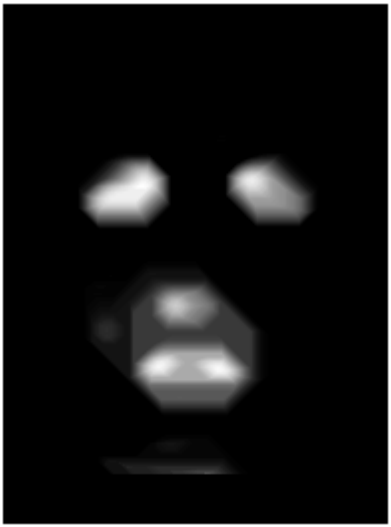 | | 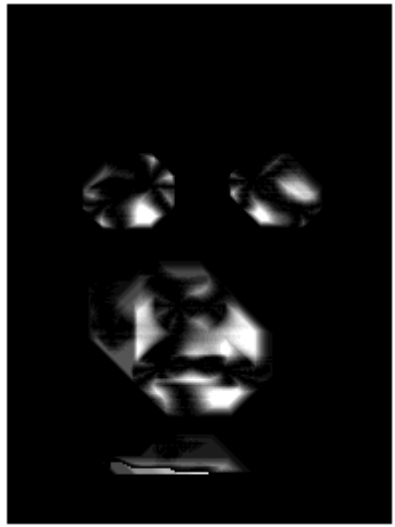 | 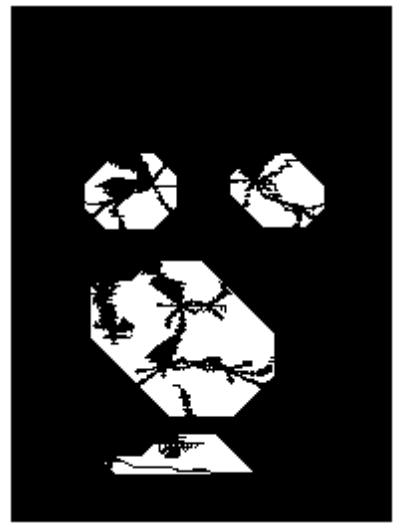 |

17.2.11. Filling the dales opened down with _expand_1234_1236() and then cleaning them with _clean_46()

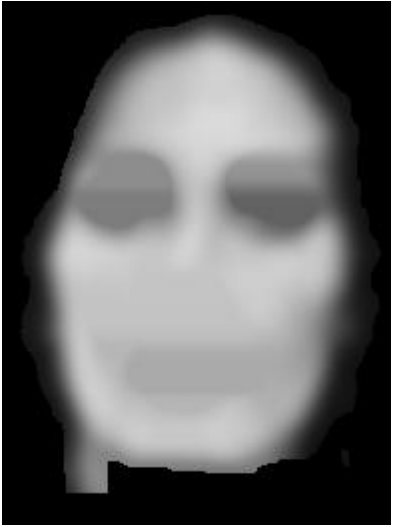

17.2.12.
17.2.11 minus 17.2.3 amplified

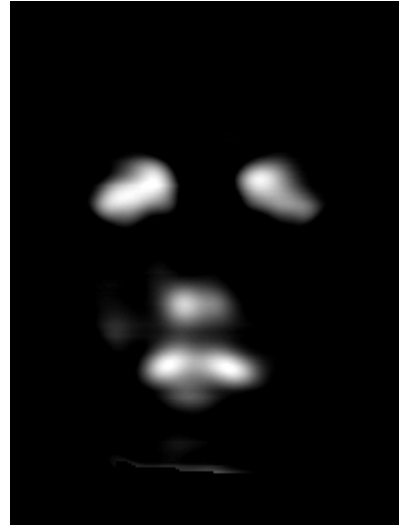

17.2.13.
XOR of 17.2.11 and 17.2.3

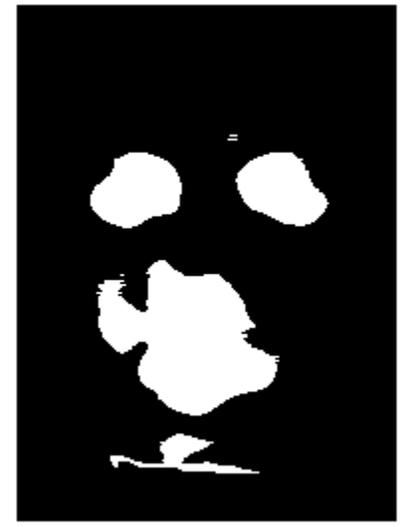

17.2.14. Filling the dales opened right with _expand_1247_1478() and then cleaning them with _clean_28()

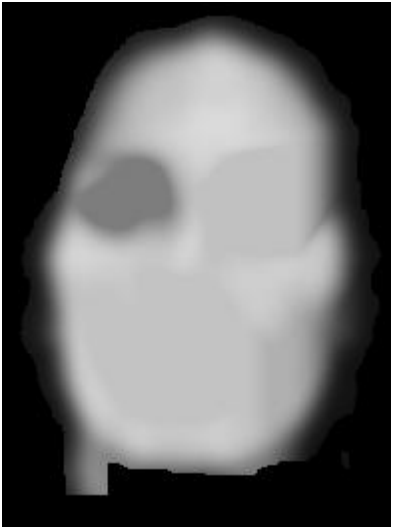

17.2.15.
17.2.14 minus 17.2.3

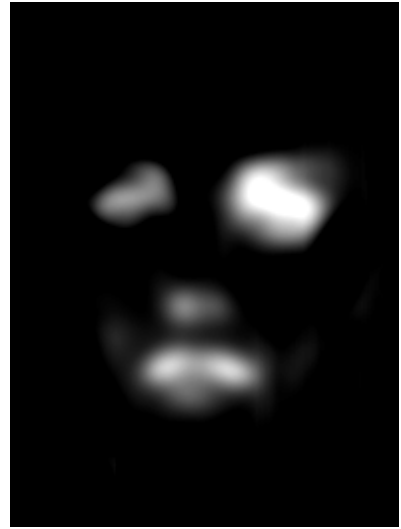

17.2.16.
XOR of 17.2.15 and 17.2.3

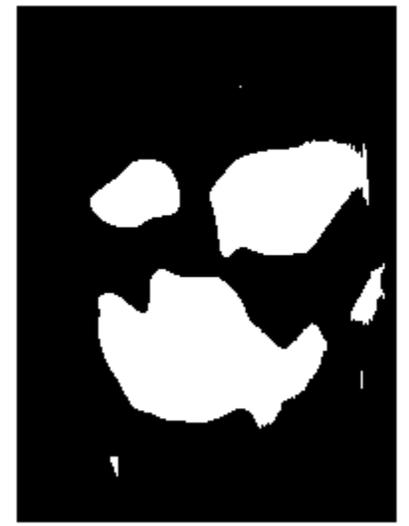

17.2.17. Filling the dales opened left with _expand_2369_3689 and then cleaning them with _clean_28()

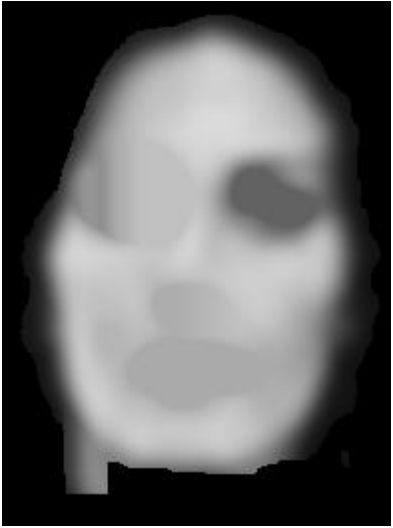

17.2.18.
17. 2.17 minus 17.2.3

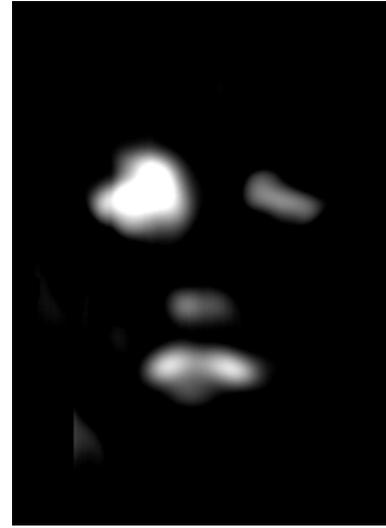

17.2.19.
XOR of 17.2.18 and 17.2.3

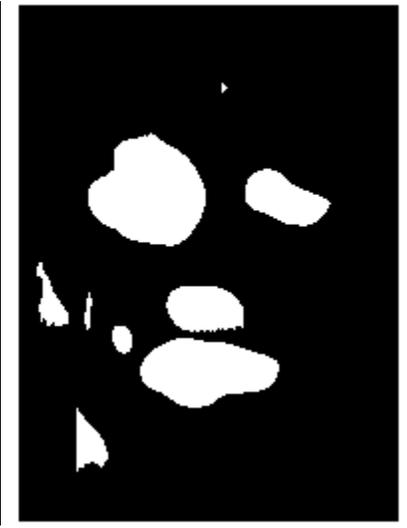

## 18. Conclusion

The quadrilateral tiling operations on ordered sets form a very large body of various operations capable of decomposing sets and images in a very comprehensive manner. An algebra is intuitively felt here. More mathematical investigation required, but as it can be seen from the examples, the operations could be used in image analysis and other areas of practice.

The operations, since their close resemblance to the structure of the living organisms neural system, non-numeric, parallel, and asynchronous nature are candidates for investigation of how living organisms process images.

## 19. "Asynchwave" Software

All the operations presented where are available for free download as the software C++ library "Asynchwave" [3]. The software includes the large testing functions of all operations, demo functions, functions which serve as documentation, and a function which produces all the image results presented in this paper.

Ansynchwave is a template C++ library produced in Microsoft Visual C++ 2012. It uses OpenCV to load images. Besides operations described here, it contains a large variety of extendable classes and functions for general purpose image analysis on image segments of arbitrary shape. The library organization is similar to STL: there are algorithms, collections, functional objects, iterators and adaptors.

\* \* \*